\documentclass[journal]{IEEEtran}
\usepackage{amsmath,amssymb}
\usepackage[ruled,norelsize]{algorithm2e}
\usepackage{multirow}
\usepackage{booktabs}
\usepackage{makecell}
\usepackage{ifpdf}
\usepackage{cite}

\ifCLASSINFOpdf
   \usepackage[pdftex]{graphicx}
  \else
  \fi

\usepackage{amsmath,amssymb}
\usepackage{array}
\usepackage{stfloats}
\makeatletter

\newcommand{\Rmnum}[1]{\expandafter\@slowromancap\romannumeral #1@}
\makeatother
\usepackage{url}
\usepackage{hyperref}
\usepackage{subfigure}
 \usepackage[usenames, dvipsnames]{color}
\newcommand{\cgreen}[1]{\textcolor{black}{#1}}
\newcommand{\cblue}[1]{\textcolor{black}{#1}}

\newcommand{\cred}[1]{\textcolor{red}{#1}}

\usepackage{ulem,url}
\newtheorem{definition}{Definition}
\usepackage{nomencl}
\makenomenclature
\hypersetup{colorlinks,       % color text of links and anchors,
            linkcolor=black,  % color for normal internal links
            anchorcolor=black,% color for anchor text
            citecolor=black,  % color for bibliographical citations
            filecolor=black,  % color for URLs which open local files
            menucolor=black,  % color for Acrobat menu items
            pagecolor=black,  % color for links to other pages
            urlcolor=black,   % color for linked URLs
	          bookmarks=true,         % create PDF bookmarks
	          bookmarksopen=false,    % don't expand bookmarks
	          bookmarksnumbered=true, % number bookmarks
	       }

\begin{document}
\title{Using Low-rank Representation of Abundance Maps and Nonnegative Tensor Factorization for Hyperspectral Nonlinear Unmixing}

\author{Lianru~Gao,~\IEEEmembership{Senior~Member,~IEEE,}
%       Yashinov Aziz,~ \IEEEmembership{Student~Member,~IEEE,}
        Zhicheng~Wang,
        Lina~Zhuang,~\IEEEmembership{Member,~IEEE,}
         Haoyang~Yu,~\IEEEmembership{Member,~IEEE,}
        Bing Zhang,~\IEEEmembership{Fellow,~IEEE,}        
        and~Jocelyn Chanussot,~\IEEEmembership{Fellow,~IEEE}% <-this % stops a space
\thanks{This work was supported in part by the National Natural Science
Foundation of China under Grant 42030111,  Grant 41722108, and Grant 42001287 and in part
 by the AXA Research Fund. \textit{(Corresponding author: Lina Zhuang.)}}
\thanks{L.  Gao  is  with  the  Key  Laboratory  of  Digital  Earth  Science, Aerospace Information Research Institute, Chinese Academy of Sciences, Beijing 100094, China (e-mail: gaolr@aircas.ac.cn).
}% <-this % stops a space
\thanks{Z. Wang and B. Zhang are with the Key Laboratory of Digital Earth Science, Aerospace Information Research Institute, Chinese Academy of Sciences, Beijing 100094, China, and also with the University of Chinese Academy of Sciences, Beijing 100049, China (e-mail: wangzc@radi.ac.cn; zb@radi.ac.cn).}
\thanks{L. Zhuang is with the Department of Mathematics, Hong Kong Baptist University, Hong Kong (e-mail: linazhuang@qq.com).}
\thanks{H. Yu is with the Center of Hyperspectral Imaging in Remote Sensing (CHIRS), Information Science and Technology College, Dalian Maritime University, Dalian, 116026, China (e-mail: yuhy@dlmu.edu.cn).}
% <-this % stops a space
\thanks{Jocelyn Chanussot is with the Universit{\'e} Grenoble Alpes, INRIA, CNRS,
Grenoble INP, LJK, 38000 Grenoble, France, and also with Aerospace Information Research Institute, Chinese Academy of Sciences, Beijing 100094,
China (e-mail: jocelyn@hi.is).}
}% <-this % stops a space


\maketitle

\begin{abstract}
Tensor-based methods have been widely studied to attack inverse problems in hyperspectral imaging, since a hyperspectral image cube can be naturally represented as a third-order tensor, which can perfectly retain the spatial information in the image. 
In this paper, we extend the linear tensor method to the nonlinear tensor method, and propose a nonlinear low-rank tensor unmixing algorithm to solve the  generalized bilinear model (GBM). Specifically, the linear and nonlinear parts of the GBM can both be expressed as   tensors. Furthermore, the low-rank structures of  
  abundance maps and nonlinear interaction abundance maps are exploited by minimizing their nuclear norm, thus taking full advantage of the high spatial correlation in hyperspectral images.
%considering the low-rank representation of abundance, the nuclear norm is added to the 
Synthetic and real-data experiments show that the low rank of abundance maps and nonlinear interaction abundance maps exploited in our method can improve the  performance of the nonlinear unmixing.
\cred{ A  MATLAB  demo  of  this  work  will be  available at  \url{https://github.com/LinaZhuang} for  the sake  of  reproducibility.}
\end{abstract}


\IEEEpeerreviewmaketitle

\section{Introduction}

\IEEEPARstart{H}{yperspectral} 
remote sensing imaging technology has developed rapidly in the past 20 years and is playing a key role in Earth observation. Hyperspectral images (HSIs) are acquired by hyperspectral cameras (HSCs) that have hundreds or thousands of spectral bands and a rich spectral resolution  ranging from $0.4\sim2.5\mu m$\cite{Keshava2002, BioucasDias2012, Dobigeon2014}.  Because of the low spatial resolution of HSCs, microscopic material mixing, and the multiple scattering that occurs within scenes\cite{BioucasDias2012}, the spectral signals in HSIs consist of mixtures of spectra of different materials. However, one benefit of the high spectral resolution of HSIs is that the problem of mixed pixels can potentially be solved. Hyperspectral unmixing (HU) aims to obtain the basic components of the image, called endmembers{,} and their corresponding proportions{,} called abundances\cite{Keshava2002}. To solve this problem, there are two main approaches that can be used. The linear mixing model (LMM) assumes that photons interact with only one material before reaching the sensor. \cblue{This  linear model  has been widely developed and tested in recent years \cite{Dobigeon2014,Zhang2014,Hong2019,Yao2019,Yu2020}, but is only suitable for   relatively simple  scenarios.} The second main method,    hyperspectral nonlinear unmixing (HNU), consists of various nonlinear mixing models (NLMMs) that consider the multiple or infinite reflections of photons. This method has been shown to be more appropriate for application to real scenes. This article focuses on HNU.

The NLMMs can be divided into two main categories: bilinear mixing models (BMMs) and high-order mixing models.
The former group includes, for example, the Fan model\cite{Fan2009}, the generalized bilinear model (GBM)\cite{Halimi2011}, \cblue{ and the linear-quadratic mixing model (LQM)\cite{Meganem2014}.}
The latter group includes the multilinear mixing model (MLM)\cite{Heylen2016}, the \textit{p}-linear model\cite{Marinoni2018}, multiharmonic postnonlinear mixing model (MHPNMM) \cite{Tang2018}, etc. 
The GBM is one of the most popular of the BMMs since it is more suitable than the Fan model (FM) for modeling scenes, and it is also a generalization of both the LMM and FM\cite{Dobigeon2014}. Meanwhile,  the GBM can more flexibly quantify the strength
of the contributions of different bilinear components than  \cblue{the polynomial post nonlinear model (PPNM) \cite{Dobigeon2014,Qu2014}}.
In this work, we address the spectral unmixing problem based on  the GBM.

The Bayesian and gradient descent algorithm (GDA) are widely used to solve optimization problems. They have been used to solve the GBM in\cite{Halimi2011a}: in this case the image was unmixed pixel by pixel and similar results were obtained. To deal with the high computational cost of pixel-based unmixing, the semi-nonnegative matrix factorization (semi-NMF) algorithm has been proposed for accelerating the optimization of  whole images in matrix form\cite{Yokoya2014}. A bound projected optimal gradient method  that transforms the GBM into a least-squares problem was also proposed in\cite{Li2016}. In addition, a multi-task learning (MTL) method that assumes that the linear part of the GBM is the task to be solved and the nonlinear part is a secondary task was proposed in \cite{Su2019}. However, all of these methods  are based on the hyperspectral image matrix, which means that the original spatial structure is destroyed and  spatial information is lost. As the HSI can be naturally expressed as a third-order tensor, which combines spectral and spatial information naturally,  tensor decomposition-based methods have been applied to  hyperspectral image analysis using canonical polyadic (CP) decomposition\cite{Veganzones2016}, Tucker decomposition\cite{Renard2008}, and  other methods.

The low rank of abundances has been exploited to solve the spectral unmixing problem  \cite{Imbiriba2018,Xiong2019}.
%It has been well documented that the low-rank of abundance maps have shown good results in linear unmixing.  
Abundances  are organized as a two-dimensional matrix ${\bf A} = [{\bf a}_1, \dots, {\bf a}_N] \in \mathbb{R}^{R \times N}$, whose columns contain $N$ abundance vectors  corresponding to $N$ pixels and rows which are abundance maps corresponding to $R$ endmembers (or atoms in a dictionary). 
\cblue{In order to  take advantage of the correlation among abundances, the
low-rank structure of abundance matrix ${\bf A}$  was exploited in \cite{Giampouras2016,Xu2016,Yang2016,Tsinos2017,Hong2018,Feng2019,Huang2019,Qu2014,Su2019}, }by minimizing the nuclear norm of matrix $ {\bf A} $ \cite{Giampouras2016,Huang2019} in optimization problems.

Apart from investigating the low-rank  abundance matrix, the low-rank structure of abundance maps was first discussed in \cite{Qian2017} in relation to solving the linear-mixing problem.
Abundances corresponding to a single endmember were organized as an abundance map, which is  a  two-dimensional matrix of size $n_{row}\times n_{col}$ (where $n_{row}$ and $n_{col}$ are the number of rows and columns in the HSI). Due to the high spatial correlation in the HSIs and the sparse distribution of endmembers, abundance maps can be approximated using  low-rank representations. In \cite{Qian2017} the low rank of abundance maps was enforced by matrix factorization: i.e., the abundance map was represented as matrix that was the product of two smaller matrices and whose size determined  the rank of the abundance map.   
Instead of using matrix factorization, we intend to enforce the low rank of abundance maps by minimizing the nuclear norm of abundance maps, which does not require prior knowledge about the rank of the abundance maps.

This paper addresses the nonlinear unmixing problem by taking advantage of the low rank of the abundance maps and nonlinear interaction maps.   
We introduce a new nonlinear unmixing method based on a tensor decomposition algorithm and low-rank representation to solve the GBM. The main contributions of this work can be summarized as follows.
\begin{itemize}
\item  \cblue{We proposed a new unmixing method based on the  GBM model. Instead of using a matrix format, GBM is re-written in a tensor format, which is a good representation for   embedding the inherent spectral-spatial structure of HSI. The tensor format allows us to address the nonlinear unmixing problem as a nonnegative tensor factorization problem and to exploit the spatial correlation of abundance maps.}
\item  The idea of the low-rank representation of abundance maps is extended to the nonlinear mixing components  - i.e., the  low-rank representation of nonlinear interaction maps. The spatial pattern of nonlinear mixing components is discussed in detail in Section II-C-2.
\end{itemize}

 The rest of the paper is organized as follows. Section \Rmnum{2} introduces the GBM model  and the proposed tensor-based algorithm.  Section \Rmnum{3} and Section \Rmnum{4} report results for  the synthetic data and real dataset, respectively. Section \Rmnum{5}  concludes the paper.

\section{Problem Formulation and Method}
In this section, we first introduce the generalized bilinear model (GBM). The proposed nonnegative tensor factorization-based method for estimating abundances is then described.
%described to solve the GBM with the tensor-based framework,which can be seen in Fig.1.

\subsection{Notation and Definitions}
In this subsection, we introduce the notation and definitions used in the paper. An \textit{n}th-order tensor is identified using Euler-cript letters – e.g.,  $\mathcal{Q}\in{\mathbb{R}^{I_{1} \times I_{2}\times ...\times I_{n}\times ...\times I_{N}}}$, with the $I_{n}$ is the size of the corresponding dimension \textit{n}. Hence, an HSI can be naturally represented as a third-order tensor,  $\mathcal{T}\in{\mathbb{R}^{I_{1} \times I_{2}\times I_{3}}}$, which consists of  $I_{1}\times I_{2}$ pixels  and $I_{3}$ spectral bands. Three further  definitions related to tensors are given as follows.

\begin{definition}
The dimension of a tensor is called the mode:  $\mathcal{Q}\in{\mathbb{R}^{I_{1} \times I_{2}\times ...\times I_{n}\times ...\times I_{N}}}$  has \textit{N} modes. For a third-order tensor $\mathcal{T}\in{\mathbb{R}^{I_{1} \times I_{2}\times I_{3}}}$, by fixing one mode, we can  obtain the corresponding sub-arrays, called slices - e.g., $\mathcal{T}_{:,:,n}$.
\end{definition}

%(Mode-$n$ product: tensor matrix multiplication) 
\begin{definition}
The mode-$n$ product of a tensor ${\cal Q} \in \mathbb{R}^{I_1 \times I_2 \times \dots \times I_N}$ by a matrix ${\bf X} \in \mathbb{R}^{J_n \times I_n}$ is a tensor ${\cal G} \in \mathbb{R}^{I_1 \times   \dots \times I_{n-1} \times J_n \times I_{n+1} \times \dots \times I_N}$, denoted as 
\begin{equation}
\nonumber
{\cal G} = {\cal Q}\times_n {\bf X},
\end{equation}
where each entry of ${\cal G}$ is defined as the sum of products of corresponding entries in ${\cal Q}$ and ${\bf X}$:
\begin{equation}
\nonumber
{\cal G}(i_1, \dots, i_{n-1}, j_n i_{n+1}, \dots, i_N) = 
\sum_{i_n} {\cal Q}(i_1, \dots, i_N) \cdot {\bf X}(j_n, i_n).
\end{equation}
\end{definition}

\begin{definition}
Given a matrix  ${\bf A} \in \mathbb{R}^{k_1 \times k_2}$ and vector ${\bf c} \in \mathbb{R}^{l_1}$, their outer product, denoted as $\textbf{A}\circ\textbf{c}$, is a tensor with dimensions $(k_{1}, k_{2}, l_{1})$ and entries
 $(\textbf{A}\circ\textbf{c})_{i1,i2,j1}=\textbf{A}_{i1,i2}\textbf{c}_{j1}$.
 \end{definition}

\subsection{Spectral Nonlinear Model:   GBM  }
Bilinear mixture models (BMMs) are based on considerations of the second-order interactions between different endmembers. These models can overcome the inherent limits of the linear model and can extract complex information from the scene to improve the unmixing results. By considering bilinear interactions as additional endmembers, a pixel ${\bf{y}} \in {\mathbb{R}^{L \times 1}}$ with $L$ spectral bands  can be expressed  as follows:
\begin{equation}
	\textbf{y} = \textbf{Ca} + \sum\limits_{i=1}^{R-1}\sum\limits_{j=i+1}^{R}\textit{b}_{i,j}\textbf{c}_i\odot\textbf{c}_j + \textbf{n},
\end{equation}
where 
 $\textbf{C}= [\textbf{c}_1, \textbf{c}_2, ..., \textbf{c}_R] \in {\mathbb{R}^{L \times R}}$, 
${\textbf{a}}= [{a}_1, {a}_2, ..., {a}_R]^\mathrm{\textit{T}} \in {\mathbb{R}^{R \times 1}}$, and
${\textbf{n}} \in {\mathbb{R}^{L \times 1}}$ 
represent the mixing matrix containing the spectral signatures of $R$ endmembers,  the fractional abundance vector, and the white Gaussian noise, respectively. 
The nonlinear coefficient ${\textit{b}_{i,j}}$  controls the nonlinear interaction between   the materials, and $\odot$ is a Hadamard product operation.

To satisfy the physical assumptions and overcome the limitations  of the FM
\cite{Dobigeon2014}, the GBM redefines the parameter $\textit{b}_{i,j}$ as $\textit{b}_{i,j}=\gamma_{i,j}\textit{a}_i\textit{a}_j$. Meanwhile, the abundance non-negativity
constraint (ANC) and the abundance sum-to-one constraint (ASC) are satisfied as follows:
\begin{align}
\nonumber {a_i} \geq 0,\sum\limits_{i = 1}^R {{a_i}}  = 1,\\
0 < {\gamma _{i,j}} < 1, \forall i < j,\\
\nonumber {\gamma _{i,j}} = 0,\forall i \geq j.
\end{align}

 The spectral mixing model for $N$ pixels can be written in matrix as:   
\begin{equation}
\label{eq:NMF}
 \textbf{Y} = \textbf{CA} + \textbf{MB} + \textbf{N},     
\end{equation}
 where ${\textbf{Y}}=[\textbf{y}_1, \textbf{y}_2, ..., \textbf{y}_N] \in {\mathbb{R}^{L \times N}}$,
${\textbf{A}}= [{\textbf{a}}_{1}, {\textbf{a}}_{2}, ... {\textbf{a}}_{N}] \in {\mathbb{R}^{R \times N}}$,
${\textbf{M}} \in {\mathbb{R}^{L \times {R(R - 1)/2}}}$,
${\textbf{B}} \in {\mathbb{R}^{{R(R - 1)/2} \times N}}$, and
 ${\textbf{N}} \in {\mathbb{R}^{L \times N}}$
 represent
  the hyperspectral image matrix,   the fractional abundance matrix with $N$ abundance vectors (the columns of ${\bf A}$),  the bilinear interaction endmember matrix,   the nonlinear interaction abundance matrix, and   the white Gaussian noise matrix, respectively.

This work aims to solve \cblue{a model-based }supervised unmixing problem: that is to estimate the abundances, ${\bf A}$, and nonlinear coefficients, ${\bf B}$, given the spectral signatures of the endmembers, \textbf{C}, which are known \cblue{as a prior}.

%\IEEEpubidadjcol
\subsection{Nonlinear Unmixing Based on Nonnegative Tensor Factorization}
\subsubsection{Tensor Framework  of  the GBM}
Traditional nonlinear unmixing methods, such as the GDA  and semi-NMF,  transform the hyperspectral image cube into a two-dimensional matrix for processing, thus destroying the internal spatial structure of the data and resulting in poor abundance inversion. However,  given that the hyperspectral images can be naturally represented as a third-order tensor,  we propose a hyperspectral nonlinear unmixing algorithm based on tensor representation for the original hyperspectral image cube. The hyperspectral image cube ${\mathcal{Y}} \in {\mathbb{R}^{n_{row} \times n_{col}\times L}}$ can be expressed in the following format:

\begin{equation}
\label{eq:mixing_tensor}
	\mathcal{Y}  = \mathcal{A}\times_{3}
	\textbf{C} + \mathcal{B}\times_{3}
	\textbf{M} + \mathcal{N},
\end{equation}
where ${\mathcal{A} \in {\mathbb{R}^{n_{row} \times n_{col} \times R}}}$,
 ${\mathcal{B} \in {\mathbb{R}^{n_{row} \times n_{col} \times {R(R - 1)/2}}}}$, and
${\mathcal{N}} \in {\mathbb{R}^{n_{row} \times n_{col}\times L}}$
denote
  the abundance cube   containing \textit{R} endmembers,
  the nonlinear interaction abundance cube,  and the white Gaussian noise cube, respectively.
  
   \cblue{The tensor model (\ref{eq:mixing_tensor}) is equivalent to the matrix model (\ref{eq:NMF}), however,  using a tensor form, enables us to remain the spatial
structure of abundances in the model, so that its spatial structure can be exploited by adding regularizations.}

\begin{figure*}[htbp]
\centering
\includegraphics[ scale=0.25]{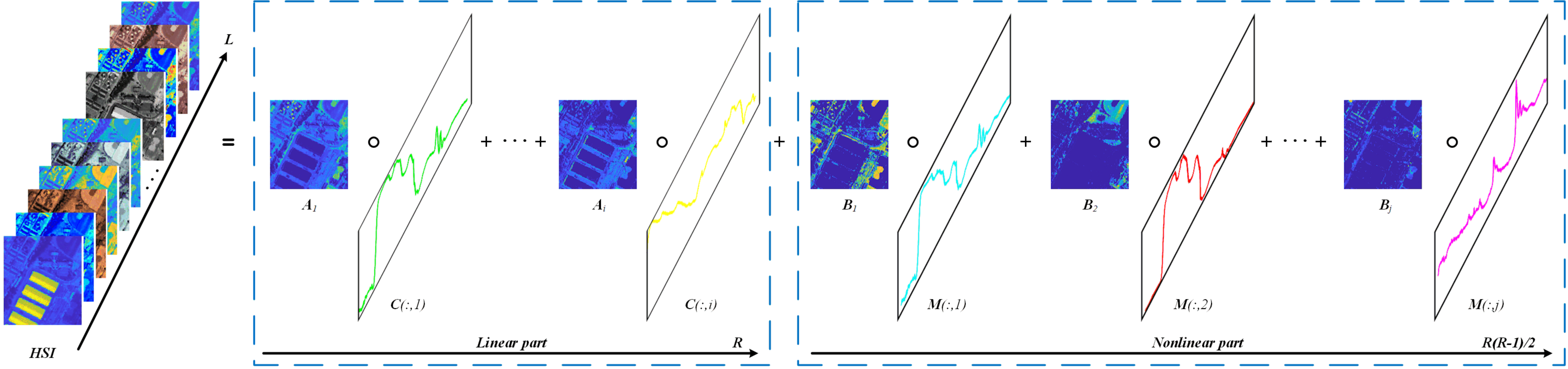}
\caption{The generalized bilinear model using the tensor-based framework}
\label{fig:GBM-framework}
\end{figure*}

To better represent the structure of abundance maps, mixing model (\ref{eq:mixing_tensor}) can be equivalently written as
\begin{equation}
\label{eq:mixing_additive}
	\mathcal{Y}  = \sum\limits_{i = 1}^R {\mathcal{A}_{:,:,i}}  \circ {\textbf{c}_i} + \sum\limits_{j = 1}^{R(R - 1)/2} {{\mathcal{B}_{:,:,j}}}  \circ {\textbf{m}_j} + \mathcal{N},
\end{equation}
where ${\mathcal{A}_{:,:,i}} \in {\mathbb{R}^{n_{row} \times n_{col}}}$,
${\textbf{c}_i} \in {\mathbb{R}^{L \times 1}}$,
 ${\mathcal{B}_{:,:,j}} \in {\mathbb{R}^{n_{row} \times n_{col}}}$, and
  ${\textbf{m}_j} \in {\mathbb{R}^{L \times 1}}$ 
 denote
the $\textit{i}$th abundance slice,   the $\textit{i}$th endmember vector,  the $\textit{j}$th interaction abundance slice, and  the $\textit{j}$th interaction endmember vector, respectively.
 Model (\ref{eq:mixing_additive}) is depicted  in Fig. \ref{fig:GBM-framework}.

\subsubsection{ \cblue{Low-rankness of Interaction Abundance Maps} }
As geographic data, the materials in HSIs tend to be spatially dependent. Spatial dependence means that things that are spatially close together tend to be more closely related than things that are far apart. Thus, the spatial distribution of a single material/endmember tends to be aggregated instead of being purely random. \cblue{This spatial distribution of a endmember enables us to approximate its abundance map with a low-rank matrix. The abundance map of a material was represented as a matrix that was the product of two smaller matrices and whose size determined  the rank of the abundance map, which was firstly introduced for the linear unmixing model in \cite{Qian2017}}.     \cblue{In this paper, we further study the low-rank representation of interaction abundance maps.}
 We exploit this spatial pattern of endmembers using a low-rank approximation for the \cblue{interaction} abundance map.
 
The low-rank structure of abundance maps and \cblue{interaction abundance maps} can be illustrated using an example.  The nonlinear unmixing of a  Cuprite image of size $250 \times 191 $ pixels was carried out.  \cblue{Fig. \ref{fig:low-rank-map} shows the abundance maps and interaction abundance maps estimated by the GDA \cite{Halimi2011a}  followed by an endmember estimation step (which uses vertex component analysis (VCA)\cite{Nascimento2005}). To see the matrix structure of abundance maps, we performed singular
value decomposition (SVD) of each abundance map $\mathcal{A}_{:,:,i} (i=1, \dots, R)$ and each interaction abundance map $\mathcal{B}_{:,:,j} (j=1, \dots, R(R-1)/2)$. Singular values are plotted in the final column of Fig. \ref{fig:low-rank-map}, where blue curves represent the    cumulative probability distributions of singular values. When the cumulative probability curve  reaches $95\%$, the corresponding dimension is marked using a pink dashed line.}
 Take the first row of Fig. \ref{fig:low-rank-map} as an \cblue{abundance maps'} example. It can be seen that $95\%$  of the data variability can be represented well in a subspace with 126 dimensions. A low-rank representation of the abundance map of the endmember ‘Sphene’ is given in Fig. \ref{fig:low-rank-map}(b), and is a very good approximation  of the original abundance map (as implied by the difference map in Fig. \ref{fig:low-rank-map}(c)). 
Furthermore, it can be seen that for interaction abundance maps shown in Fig. \ref{fig:low-rank-map}(m) and Fig. \ref{fig:low-rank-map}(q), the cumulative probability curves reaches 95\% when the corresponding dimensions are 85 and 21, respectively, which are much smaller than the subspace dimensions in the abundance maps. 
 \cblue{The corresponding  low-rank representation of the interaction abundance maps are shown Fig. \ref{fig:low-rank-map}(n) and Fig. \ref{fig:low-rank-map}(r), which indicate that the interaction abundance maps can be approximated well by  low-rank  matrices.}
 We remark that the ranks of abundance maps is usually not as low as the rank of image matrix,  ${\bf Y}$, which usually can be approximated using a subspace with dimension lower than 20. The rank of a abundance map depends on the spatial distribution of the corresponding endmember. As long as a abundance map is not a full-rank matrix, its low-rankness can be imposed in the objective function to obtain a better estimate of the abundance map. 
The lower the rank is, the better result of the estimated abundance map we can expect. It has been demonstrated in the work \cite{Qian2017} that the use of low-rankness of abundance maps can improve linear unmixing performance. In Fig. \ref{fig:low-rank-map}, we justify that the ranks of interaction abundance maps are lower than the ranks of abundance maps, implying the use of low-rankness of interaction abundance maps can also improve nonlinear unmixing performance.

To take full advantage of the low-rank structure of abundance maps \cblue{and interaction abundance maps}, we propose a new nonlinear unmixing method based \cblue{both} on the {\textbf L}ow-{\textbf R}ank representation of  abundance maps and \cblue{interaction abundance maps} \cblue{via}  {\textbf N}onnegative {\textbf T}ensor {\textbf F}actorization, termed \textbf{LR-NTF}, which aims to solve the following optimization problem:

\begin{figure*}[htbp]
\centering
\includegraphics[scale=0.45]{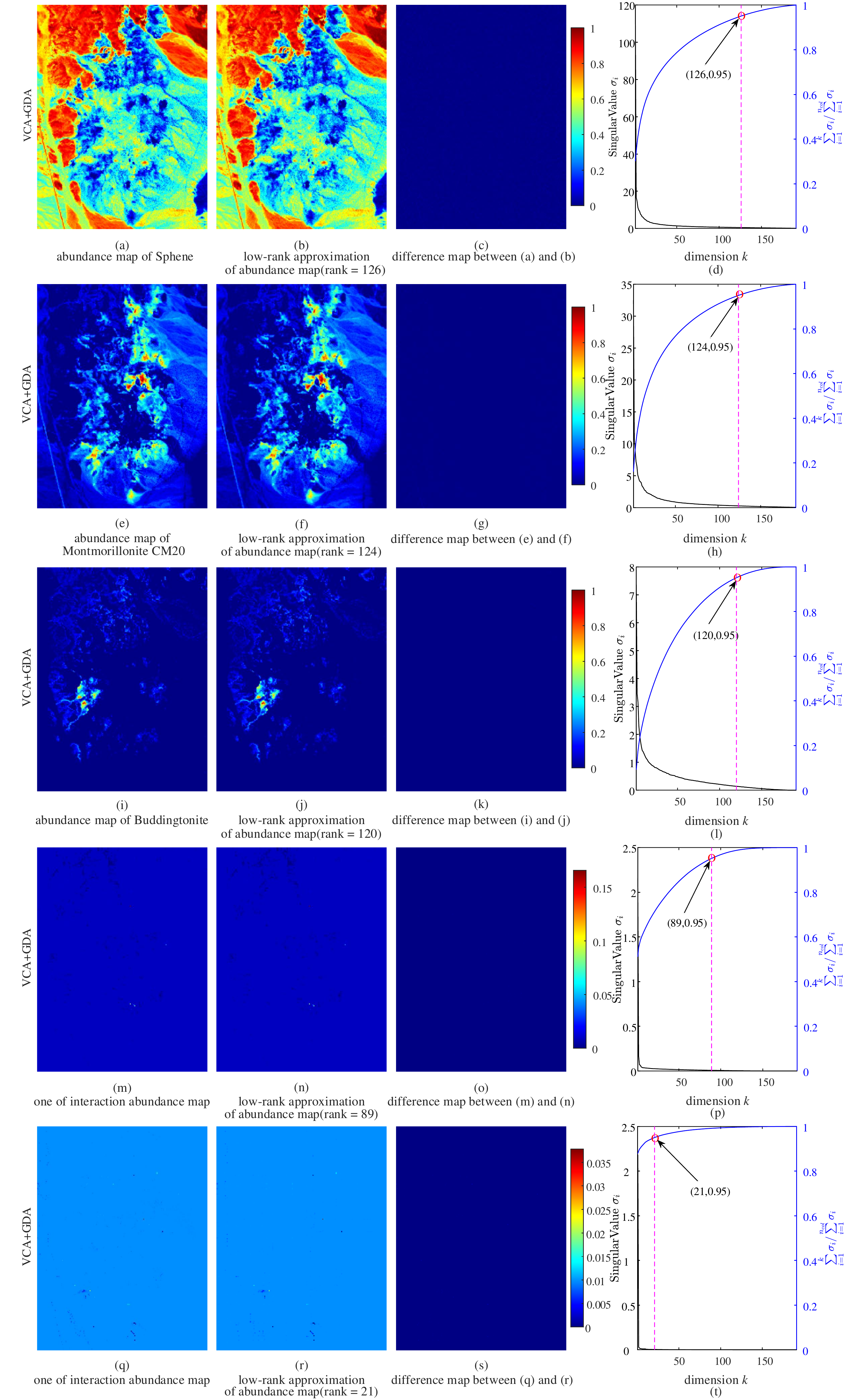}
\caption{Low-rank approximation of abundance maps and   nonlinear interaction abundance maps in the Cuprite image}
\label{fig:low-rank-map}
\end{figure*}

\begin{gather}
\setlength{\abovedisplayskip}{0pt}
\setlength{\belowdisplayskip}{0pt}
\nonumber\mathop {\mathop{\mathop {\arg\min }\limits_{{\mathcal{A}_{:,:,i}\geq0}, \mathcal{A}^*_{:,:,j} \geq {\mathcal{B}_{:,:,j}\geq0}}}\limits_{i = 1,2,...,R,}}\limits_{j = 1,2,...,R(R - 1)/2} \frac{{\rm{1}}}{{\rm{2}}}\left\| {\mathcal{Y}  \!-\! \sum\limits_{i = 1}^R {{\mathcal{A}_{:,:,i}}} \! \circ \!{\textbf{c}_i} \!-\! \sum\limits_{j = 1}^{R(R - 1)/2} {{\mathcal{B}_{:,:,j}}} \! \circ \! {\textbf{m}_j}} \right\|_F^2 \\
\nonumber+ {\lambda _1}(\sum\limits_{i = 1}^R {{{\left\| {{\mathcal{A}_{:,:,i}}} \right\|}_ * })}  + {\lambda _2}(\sum\limits_{j = 1}^{R(R - 1)/2} {{{\left\| {{\mathcal{B}_{:,:,j}}} \right\|}_ * })} \\
s.t. \sum\limits_{i = 1}^R {{\mathcal{A}_{:,:,i}}} {\kern 1pt} {\kern 1pt}  = {{\textbf{1}}_{{n_{row}}}}{\textbf{1}}_{{n_{col}}}^T,
\label{eq:optimi1}
\end{gather}
where
$\mathcal{A}^*_{k_1,k_2,(i,j)} = \mathcal{A}_{k_1,k_2,i}\mathcal{A}_{k_1,k_2,j} (k_1 \in \{1, \dots, n_{row}\}, k_2 \in \{1, \dots, n_{col}\}, ~\text{and}~ (i,j) \in \{(i,j)| i<j, i\in \{1, \dots, R \}, j \in \{ 1, \dots, R \}\} ) $, and
 ${{{\left\| {\mathcal{X} } \right\|}_F^2}}$ denotes the Frobenius norm which returns the square root of the sum of the absolute squares of its elements.
The symbol ${{{\left\| {{\cdot}} \right\|}_ * }}$ represents the nuclear norm, and $\mathbf{1}_{d}$ represents a vector whose components are all one and whose dimension is given by its subscript.  Let $\textbf{P} =\mathbf{1}_{n_{row}} \mathbf{1}_{n_{col}}^T$. Abundance maps $\mathcal{A}_{:,:,i}$ and $\mathcal{B}_{:,:,j}$ are enforced to be low-rank by minimizing their nuclear norms.  
$\lambda_1 > 0$ and $\lambda_2  > 0 $ are parameters of regularizations.

The optimization problem in (\ref{eq:optimi1}) can be solved by
optimization using {alternating direction method of multipliers (ADMM)} \cite{1992On}. To use the ADMM,  first  (\ref{eq:optimi1}) is converted into an equivalent form by introducing multiple auxiliary variables $\textbf{V}_i$, $\textbf{E}_j$ to replace $\mathcal{A}_{:,:,i}(i=1, ..., R)$, $\mathcal{B}_{:,:,j}(j=1, ..., R(R-1)/2)$. The formulation is as follows:

\begin{gather}
\nonumber\mathop {\mathop{\mathop {\arg\min }\limits_{{\mathcal{A}_{:,:,i}\geq0}, \mathcal{A}^*_{:,:,j} \geq {\mathcal{B}_{:,:,j}\geq0}}}\limits_{i = 1,2,...,R,}}\limits_{j = 1,2,...,R(R - 1)/2}\frac{{\rm{1}}}{{\rm{2}}}\left\| {\mathcal{Y} \! -\! \sum\limits_{i = 1}^R {{\mathcal{A}_{:,:,i}}} \! \circ \!{\textbf{c}_i}\! - \!\sum\limits_{j = 1}^{R(R - 1)/2} {{\mathcal{B}_{:,:,j}}} \! \circ\! {\textbf{m}_j}} \right\|_F^2 \\+ \nonumber{\lambda _1}(\sum\limits_{i = 1}^R {{{\left\| {{\textbf{V}_i}} \right\|}_ * })}  + {\lambda _2}(\sum\limits_{j = 1}^{R(R - 1)/2} {{{\left\| {{\textbf{E}_j}} \right\|}_ * })} \\
s.t. \left\{ {\begin{array}{*{20}{c}}
{{\mathcal{A}_{:,:,i}} = {\textbf{V}_i},i = 1,2,...,R}\\
{{\mathcal{B}_{:,:,j}} = {\textbf{E}_j},j = 1,2,...,R(R - 1)/2}\\
{\sum\limits_{i = 1}^R {{\mathcal{A}_{:,:,i}} = \mathbf{1}_{n_{row}} \mathbf{1}_{n_{col}}^T} }
\end{array}} \right. .
\label{eq:optimi2}
\end{gather}

 By using the  Lagrangian function,   (\ref{eq:optimi2}) can be reformulated as:
\begin{equation}
\begin{aligned}
&  \mathcal{L}({\mathcal{A}_{:,:,i}},{\mathcal{B}_{:,:,j}},{\textbf{V}_i},{\textbf{E}_j},{\textbf{D}_i},{\textbf{H}_j},\textbf{G})=\\
& \frac{{\rm{1}}}{{\rm{2}}}\left\| {\mathcal{Y}  - \sum\limits_{i = 1}^R {{\mathcal{A}_{:,:,i}}}  \circ {\textbf{c}_i} - \sum\limits_{j = 1}^{R(R - 1)/2} {{\mathcal{B}_{:,:,j}}}  \circ {\textbf{m}_j}} \right\|_F^2 + {\lambda _1}(\sum\limits_{i = 1}^R {{{\left\| {{\textbf{V}_i}} \right\|}_ * })} \\
&  + {\lambda _2}(\sum\limits_{j = 1}^{R(R - 1)/2} {{{\left\| {{\textbf{E}_j}} \right\|}_ * })}  + \frac{\mu }{2}(\sum\limits_{i = 1}^R {\left\| {{\mathcal{A}_{:,:,i}} - {\textbf{V}_i} - {\textbf{D}_i}} \right\|_F^2)}    +  \\
&  \frac{\mu }{2}(\sum\limits_{j = 1}^{R(R - 1)/2} {\left\| {{\mathcal{B}_{:,:,j}} - {\textbf{E}_j} - {\textbf{H}_j}} \right\|_F^2)}+\frac{\mu }{2}\left\| {\sum\limits_{i = 1}^R {{\mathcal{A}_{:,:,i}}}  - {\textbf{P}} - \textbf{G}} \right\|_F^2,
\end{aligned}
\label{eq:optimi3}
\end{equation}
where  $\textbf{D}_i$, $\textbf{H}_j$ and $\textbf{G}$ are Lagrange multipliers, and   $\mu$ is the penalty parameter. The variables ${\mathcal{A}_{:,:,i}},{\mathcal{B}_{:,:,j}}, {\textbf{V}_i}, {\textbf{E}_j}, {\textbf{D}_i}, {\textbf{H}_j}, \textbf{G}$ were updated sequentially: this step is shown in \textbf{Algorithm 1}. The optimization details of the loss function are as given in the following:

Update  $\mathcal{A}$. The optimization problem
for $\mathcal{A}_{:,:,i}$ is
\begin{equation}
\begin{aligned}
& \mathcal{A}_{:,:,i}^{k + 1}  = \mathop {\arg\min}\limits_{\mathcal{A}_{:,:,i}^k} \frac{{\rm{1}}}{{\rm{2}}}\left\| {\mathcal{Y}  - \sum\limits_{i = 1}^R {\mathcal{A}_{:,:,i}^k}  \circ {\textbf{c}_i} - \sum\limits_{j = 1}^{R(R - 1)/2} {\mathcal{B}_{:,:,j}^k}  \circ {\textbf{m}_j}} \right\|_F^2 \\
 &+ \frac{\mu }{2}\left\| {\mathcal{A}_{:,:,i}^k - \textbf{V}_i^k - \textbf{D}_i^k} \right\|_F^2 + \frac{\mu }{2}\left\| {\sum\limits_{i = 1}^R {\mathcal{A}_{:,:,i}^k}  - {\textbf{P}} - {\textbf{G}^k}} \right\|_F^2 \\ 
& = \cblue{ \mathop {\arg\min}\limits_{\mathcal{A}_{:,:,i}^k}}\frac{{\rm{1}}}{{\rm{2}}}\sum\limits_{b = 1}^L {\left\| {{\mathcal{O}_{:,:,b}} - \mathcal{A}_{:,:,i}^k{\textit{c}_{{b_i}}}} \right\|_F^2}  + \frac{\mu }{2}(\sum\limits_{i = 1}^R {\left\| {\mathcal{A}_{:,:,i}^k - \textbf{V}_i^k - \textbf{D}_i^k} \right\|_F^2)} \\
&+ \frac{\mu }{2}\left\| {\mathcal{A}_{:,:,i}^k + \widetilde {\bf{\textbf{A}}} - {\textbf{P}} - {\textbf{G}^k}} \right\|_F^2,
\end{aligned}
\end{equation}
where  $\mathcal{O}=\mathcal{Y}-\sum\limits_{i = 1,\neg i}^R {\mathcal{A}_{:,:,i}^k}  \circ {\textbf{c}_i} - \sum\limits_{j = 1}^{R(R - 1)/2} {\mathcal{B}_{:,:,j}^k}  \circ {\textbf{m}_j} \in {\mathbb{R}^{n_{row} \times n_{col}\times L}}$, and  $\mathcal{O}_{:,:,b}$ is the \textit{b}th slice. Meanwhile, $\widetilde {\bf{\textbf{A}}} = {\sum\limits_{i = 1,\neg i}^R {\mathcal{A}_{:,:,i}^k}\in {\mathbb{R}^{n_{row} \times n_{col}}}}$ \cblue{represents the sum of all abundance maps apart from  the \textit{i}th slice,
$\sum\limits_{i = 1,\neg i}^R a_i = a_1 + \dots + a_{i-1} + a_{i+1} + \dots + a_R$, 
 and } $\textbf{c}_{i}=[\textit{c}_{{i_1}},\textit{c}_{{i_2}}...,\textit{c}_{{b_i}},...,\textit{c}_{{i_L}}]^\mathrm{\textit{T}} \in {\mathbb{R}^{L \times 1}}$ is the \textit{i}th endmember. Hence the solution for $\mathcal{A}_{:,:,i}$ can be derived as follows:
\begin{equation}
\begin{aligned}
%\nonumber \frac{{\partial \mathcal{L}}}{{\partial \mathcal{A}_{:,:,i}^k}} & =  - \sum\limits_{b = 1}^B {(\widetilde {{\mathcal{Y}_b}} - \textbf{A}_i^k{\textit{c}_{{b_i}}})} \textit{c}_{{b_i}}^T + \mu (\textbf{A}_i^k - \textbf{V}_i^k - \textbf{D}_i^k) + \mu (\textbf{A}_i^k + \widetilde {\bf{A}} - {\bf{1}} - {\textbf{D}^k}) = 0\\
 &\mathcal{A}_{:,:,i}^{k + 1}=\\
  & {(\sum\limits_{b = 1}^L {{\textit{c}_{{b_i}}}} \textit{c}_{{b_i}}^T + 2\mu \textbf{I})^{ - 1}}(\sum\limits_{b = 1}^B {{{\mathcal{O}_{:,:b}}}\textit{c}_{{b_i}}^T + } \mu (\textbf{V}_i^k + \textbf{D}_i^k + {\textbf{P}}{\rm{ +  }} { \textbf{G}^k}{\rm{ - }}\widetilde {\bf{A}})).
\end{aligned}
\end{equation}

Update  $\mathcal{B}$. The optimization problem
for $\mathcal{B}_{:,:,j}$ is
\begin{equation}
\begin{aligned}
\mathcal{B}_{:,:,j}^{k + 1} &= \mathop {\arg\min}\limits_{\mathcal{B}_{:,:,j}^k} \frac{{\rm{1}}}{{\rm{2}}}\left\| {\mathcal{Y}  - \sum\limits_{i = 1}^R {\mathcal{A}_{:,:,i}^{k + 1}}  \circ {\textbf{c}_i} - \sum\limits_{j = 1}^{R(R - 1)/2} {\mathcal{B}_{:,:,j}^k}  \circ {\textbf{m}_j}} \right\|_F^2\\
& + \frac{\mu }{2}\left\| {\mathcal{B}_{:,:,j}^k - \textbf{E}_j^k - \textbf{H}_j^k} \right\|_F^2\\
 & = \frac{{\rm{1}}}{{\rm{2}}}\sum\limits_{b = 1}^L {\left\| { {{\mathcal{K}_{:,:,b}}} - \mathcal{B}_{:,:,j}^k{\textit{m}_{{b_j}}}} \right\|_F^2}  + \frac{\mu }{2}\left\| {\mathcal{B}_{:,:,j}^k - \textbf{E}_j^k - \textbf{H}_j^k} \right\|_F^2,
\end{aligned}
\end{equation}
where  $\mathcal{K}=\mathcal{Y}-\sum\limits_{i = 1}^R {\mathcal{A}_{:,:,i}^k}  \circ {\textbf{c}_i} - \sum\limits_{j = 1,,\neg j}^{R(R - 1)/2} {\mathcal{B}_{:,:,j}^k}  \circ {\textbf{m}_j} \in {\mathbb{R}^{n_{row} \times n_{col}\times L}}$, and  $\mathcal{K}_{:,:,b}$ is the \textit{b}th slice. Meanwhile, $\textbf{m}_{j}=[\textit{m}_{{j_1}},\textit{m}_{{j_2}}...,\textit{m}_{{b_j}},...,\textit{m}_{{j_L}}]^\mathrm{\textit{T}} \in {\mathbb{R}^{L \times 1}}$ is the \textit{j}th interaction endmember. Hence the solution for $\mathcal{B}_{:,:,j}$ can be derived as follows:
\begin{equation}
\begin{aligned}
%\nonumber \frac{{\partial \mathcal{L}}}{{\partial \mathcal{B}_{:,:,j}^k}} &=  - \sum\limits_{b = 1}^B {(\widetilde {\widetilde {{\mathcal{Y}_b}}} - \mathcal{B}_{:,:,j}^k{\textbf{m}_{{b_j}}})} \textbf{m}_{{b_j}}^T + \mu (\textbf{B}_j^k - \textbf{E}_j^k - \textbf{D}_j^k) = 0\\
 \mathcal{B}_{:,:,j}^{k + 1} & = {(\sum\limits_{b = 1}^L {{\textit{m}_{{b_j}}}} \textit{m}_{{b_j}}^T + \mu \textbf{I})^{ - 1}}(\sum\limits_{b = 1}^L { {{\mathcal{K}_{:,:,b}}}\textit{m}_{{b_j}}^T + } \mu (\textbf{E}_j^k + \textbf{H}_j^k)).
 \end{aligned}
\end{equation}

Update  $\textbf{V}$. The optimization problem
for $\textbf{V}_{i}$ is
\begin{equation}
\begin{aligned}
\textbf{V}_i^{k + 1} & = \mathop {\arg\min}\limits_{\textbf{V}_i^k} {\lambda_1}{\left\| {\textbf{V}_i^k} \right\|_* } + \frac{\mu }{2}\left\| {\mathcal{A}_{:,:,i}^{k + 1} - \textbf{V}_i^{k} - \textbf{D}_i^k} \right\|_F^2\\
& = \frac{1}{2}\left\| {\widetilde{\textbf{V}}_{i} - \textbf{V}_i^k} \right\|_F^2 + \frac{{{\lambda_1}}}{\mu }{\left\| {\textbf{V}_i^k} \right\|_ * },
 \end{aligned}
\label{eq:optimi-V}
\end{equation}
where  $\widetilde{\textbf{V}}_{i}=\mathcal{A}_{:,:,i}^{k + 1} - \textbf{D}_i^k \in {\mathbb{R}^{n_{row} \times n_{col}}}$. Sub-problem (\ref{eq:optimi-V}) can be solved using singular value composition (SVD) of $\widetilde{\bf V}_i$ :
\begin{equation}
	[\textbf{U,S,Z}] = \text{SVD}(\widetilde{\textbf{V}}_{i}),
\end{equation}
where  \textbf{U}  and \textbf{Z} are  orthogonal matrices. Meanwhile,  $\textbf{S}$ is the singular value matrix. A soft-thresholding
operator is applied to singular values:
\begin{equation}
\widetilde{\textbf{S}}  = diag(\max (diag(\textbf{S}) - \frac{{{\lambda _1}}}{\mu },0)),
\label{eq:optimi-S}
\end{equation}
where  $diag(\textbf{.})$ returns a column vector consisting of the main diagonal elements if the input variable is a matrix and returns a diagonal matrix with the elements on the main diagonal if the input variable is a vector.  $\textbf{V}_i^{k + 1}$ can then be expressed as 
\begin{equation}
\textbf{V}_i^{k + 1} = \textbf{U}\widetilde{\textbf{S}}{\textbf{Z}^T}.
\end{equation}

Update to $\textbf{E}$. The optimization problem
for $\textbf{E}_{j}$ is
\begin{equation}
\begin{aligned}
 \textbf{E}_j^{k + 1} & = \mathop {\arg\min}\limits_{\textbf{E}_j^k} {\lambda_2}{\left\| {\textbf{E}_j^k} \right\|_ * } + \frac{\mu }{2}\left\| {\mathcal{B}_{:,:,j}^{k + 1} - \textbf{E}_j^k - \textbf{H}_j^k} \right\|_F^2 \\
& = \frac{1 }{2}\left\| {\widetilde{\textbf{E}}_{j} - \textbf{E}_j^k} \right\|_F^2 + \frac{{{\lambda _2}}}{\mu }{\left\| {\textbf{E}_j^k} \right\|_ * },
 \end{aligned}
 \label{eq:optimi-E}
\end{equation}
where  $\widetilde{\textbf{E}}_{j}=\mathcal{B}_{:,:,j}^{k + 1} - \textbf{H}_j^k \in {\mathbb{R}^{n_{row} \times n_{col}}}$. Sub-problem (\ref{eq:optimi-E}) can be solved via SVD of $\widetilde{\textbf{E}}_{j}$:
\begin{equation}
	[{\bf{U, S, Z}}] = \text{SVD}(\widetilde{\bf{E}}_j),
\end{equation}
where  $\textbf{S}$ is the singular value matrix. As for (\ref{eq:optimi-S}), a soft-thresholding operator is applied to the singular values:
\begin{equation}
\widetilde{\textbf{S}}  = diag(\max (diag(\textbf{S}) - \frac{{{\lambda _2}}}{\mu },0)),
\end{equation}
 then $\textbf{E}_j^{k + 1}$ can be expressed as 
\begin{equation}
\textbf{E}_j^{k + 1} = \textbf{U}\widetilde{\textbf{S}}{\textbf{Z}^T}.
\end{equation}

Update  $\textbf{D}_i$.
\begin{equation}
\textbf{D}_i^{k + 1} = \textbf{D}_i^k - (\mathcal{A}_{:,:,i}^{k + 1} - \textbf{V}_i^{k + 1}).
\end{equation}

Update  $\textbf{H}_j$.
\begin{equation}
\textbf{H}_j^{k + 1} = \textbf{H}_j^k - (\mathcal{B}_{:,:,j}^{k + 1} - \textbf{E}_j^{k + 1}).
\end{equation}

Update  \textbf{G}.
\begin{equation}
{\textbf{G}^{k + 1}} = {\textbf{G}^k} - (\sum\limits_{i = 1}^R {\mathcal{A}_{:,:,i}^{k + 1}}  - {\textbf{P}}).
\end{equation}

%\begin{figure}[!t]

\begin{algorithm}[h]
\normalem
 \caption{The Proposed LR-NTF Algorithm} 
 \LinesNumbered
 \KwIn{HSI: $\mathcal{Y}$; Matrix of Endmembers: $\textbf{E}$}
 \KwOut{A cube of abundance maps: $\mathcal{A}$}
 \For( ){$k=1$; $k<Iter$; $k++$}
 {
    $ \mathcal{A}_{:,:,i}^{k + 1} = {(\sum\limits_{b = 1}^L {{\textit{c}_{{b_i}}}} \textit{c}_{{b_i}}^T + 2\mu \textbf{I})^{ - 1}}(\sum\limits_{b = 1}^L { {{\mathcal{O}_{:,:b}}}\textit{c}_{{b_i}}^T + } \mu (\textbf{V}_i^k + \textbf{D}_i^k + {\textbf{P}}{\rm{ + }}{\textbf{G}^k}{\rm{ - }}\widetilde {\bf{A}}))$\;
    $\mathcal{A}=\text{abs}(\mathcal{A})$\;
    $\mathcal{B}_{:,:,j}^{k + 1}  = {(\sum\limits_{b = 1}^L {{\textit{m}_{{b_j}}}} \textit{m}_{{b_j}}^T + \mu \textbf{I})^{ - 1}}(\sum\limits_{b = 1}^L { {{\mathcal{K}_{:,:,b}}}\textit{m}_{{b_j}}^T + } \mu (\textbf{E}_j^k + \textbf{H}_j^k))$. If any element of $\mathcal{B}$ exceeds that of $\mathcal{A}^*$, it is replaced with that of $\mathcal{A}^*$\;
    $\mathcal{B}=\text{abs}(\mathcal{B})$\;
    $\textbf{V}_i^{k + 1} = \textbf{U}\widetilde{\textbf{S}}{\textbf{Z}^T}$\;
    $\textbf{E}_j^{k + 1} = \textbf{U}\widetilde{\textbf{S}}{\textbf{Z}^T}$\;
    $\textbf{D}_i^{k + 1} = \textbf{D}_i^k - (\mathcal{A}_{:,:,i}^{k + 1} - \textbf{V}_i^{k + 1})$\;
    $\textbf{H}_j^{k + 1} = \textbf{H}_j^k - (\mathcal{B}_{:,:,j}^{k + 1} - \textbf{E}_j^{k + 1})$\;
    ${\textbf{G}^{k + 1}} = {\textbf{G}^k} - (\sum\limits_{i = 1}^R {\mathcal{A}_{:,:,i}^{k + 1}}  - {\textbf{P}})$\;
    $ k = k +1 $;
 }
 \Return result
\end{algorithm}

\section{Experiments on Synthetic Data}

In this section, we compare the performance of the proposed algorithm LR-NTF and other algorithms including the GDA and semi-NMF. The GDA is considered to be the benchmark for solving the generalized bilinear model, and semi-NMF solves the unmixing problem in matrix form.
 
Two widely used metrics, namely the root-mean-square error (RMSE) of abundances and the  image reconstruction error (RE) were adopted for evaluating  the unmixing methods. The RMSE quantifies the difference between the estimated abundances $\widehat{\mathcal A}$ and the true abundances ${\mathcal A}$
as follows: 
\begin{equation}
\cblue{\textit{RMSE}=\sqrt{\frac{1}{R \times N}  \| {\mathcal A} - \widehat{\mathcal A} \|_F^2}.}
 \label{eq:RMSE}
\end{equation}
The RE measures the difference 
between the observations ${\mathcal Y}$ and  their reconstructions  $\widehat{\mathcal Y}$   as follows: 
\begin{equation}
\cblue{\textit{RE}=\sqrt{\frac{1}{N \times L}  \| {\mathcal Y} - \widehat{\mathcal Y} \|_F^2}.}
 \label{eq:RE}
\end{equation}
\subsection{Data Generation}
In this study, the simulated data were synthesized in a similar way to the data in \cite{Miao2007}, and \cite{Qian2017}. The synthesis was carried out as follows.
\begin{enumerate}
\item  Six spectral reflection signals  with 224  spectral bands ranging from 0.38 to 2.5 $\mu$m were chosen from the United States Geological Survey (USGS) digital spectral library\footnote{\url{https://www.usgs.gov/labs/spec-lab}}. Specifically, these were Carnallite, Ammonio–jarosite, Almandine, Brucite, Axinite, and Chlonte.
\item We generated an image of size $s^2\times s^2 \times L$: 
this could be     divided into small blocks of size $s\times s \times L$. %to synthesize the abundance map.
\item A randomly selected endmember was assigned to each block, and a $k \times k$ low-pass filter was then applied to generate  abundance maps  of size $s^2\times s^2 \times R$  that contained mixed pixels while still satisfying the ANC and ASC constraints.
\item To simulate a non-pure pixel scenario, we detected the abundances whose values were higher than 0.8 and replaced them with the average fraction for all the endmembers.
\item The abundance and endmember information for the image were obtained in steps 1-4. Clean HSIs were generated based on two kinds of nonlinear mixing models, namely  the generalized bilinear model and the polynomial post nonlinear model. The interaction coefficients  in the GBM were set randomly, and the interaction coefficients in the PPNM were set to 0.25.
\item To evaluate the robustness to additive noise, zero-mean Gaussian white noise was added to the clean data.
A series of noisy images with signal-to-noise ratios (SNRs) = \{15, 20, 30, 40\} dB was generated.
\end{enumerate}
\subsection{Algorithm Evaluation}
After obtaining the synthetic data, we compared different unmixing methods, including the proposed method, the GDA, and semi-NMF. \cblue{Meanwhile,  the proposed method was compared with the minimum volume nonnegative tensor factorization (MV-NTF) \cite{Qian2017}  separately on the three nonlinear datasets to see the imapct of low-rank representation of interaction abundance maps.} We first tested the effect of different SNRs  before comparing the methods. The parameter settings for the tested methods are describe in the following.

\subsubsection{Parameter Setting} 

In this experiment, we analysed the parameters used \cblue{in  all  the algorithms}. For the GDA, which was the benchmark method,  the tolerance for stopping the iterations was set to $1\times10^{-6}$. \cblue{The  parameters in the semi-NMF  and MV-NTF were hand-tuned to get the optimal performance.} The proposed  LR-NTF  has an abundance low-rank regularization parameter $\lambda_{1}$, a nonlinear iteration abundance low-rank regularization parameter $\lambda_{2}$, and a balance parameter $\mu$ that controls the convergence rate of the algorithm. We searched for the optimal value of  $\lambda_{1}$ and $\lambda_{2}$ using a grid method with   $\lambda_{1}\in \{0.1,0.2,...,1\}$ and   $\lambda_{2} \in \{0.01,0.02,...,0.1\}$. 
To test the impact of parameters $\lambda_{1}$ and $\lambda_{2}$ on the unmixing,
  we  generated a synthetic dataset with parameters $k= 9$, $s= 6$, and SNR = 30 dB.  
  In our experiments, similar results were obtained for certain ranges of parameters: i.e. $\lambda_{1}\in \{0.1\}$  and   $\lambda_{2} \in \{0.06,0.07,0.08,0.09\}$. Hence, we set the parameters $\lambda_{1}$ and $\lambda_{2}$   to within this range in the subsequent experiments.
  
\subsubsection{	Comparison between Methods Using Different Gaussian Noise Levels}

In this experiment, three images (of size $100\times100\times224$)   were generated using two types of model (GBM and PPNM):   image1 and image2, which are described in Table \ref{tab:SNR1}, were generated using the GBM and PPNM, respectively. image3 was a mixture of image1 and image2, with   half the  pixels being generated by the GBM and the rest generated by the PPNM\cite{Yokoya2014}. The first image, generated based on GBM, is used to compare the proposed algorithm to the GDA and semi-NMF method, whereas  image2 and image3 were used to test the robustness of the proposed algorithm.
 Meanwhile, white Gaussian noise was added in the simulated images:  the signal-to-noise ratio (SNR) was set to 15 dB, 20 dB, 30 dB, and 40 dB.

\begin{table}[htbp]

\setlength{\tabcolsep}{1.3mm}{
  \centering
  \caption{The Evaluate Results of the Proposed Method and State-of-the-Art Methods}
  
    \begin{tabular}{ccccccc}
    \toprule
    Scenario & SNR   & Metric & FCLS  & GDA   & Semi-NMF & LR-NTF (ours) \\
    \midrule
    \multirow{12}[2]{*}{image1} & \multirow{3}[1]{*}{15} & RMSE  & 0.0746  & 0.0627  & 0.0608  & \textbf{0.0437 } \\
          &       & RE    & 0.0983  & 0.0975  & 0.0959  & \textbf{0.0954 } \\
          &       & \cblue{Time(s)}  & \cblue{\textbf{3 }} & \cblue{1816}  & \cblue{18}    & \cblue{1022}  \\
          & \multirow{3}[0]{*}{20} & RMSE  & 0.0680  & 0.0534  & 0.0473  & \textbf{0.0253 } \\
          &       & RE    & 0.0582  & 0.0570  & 0.0542  & \textbf{0.0537 } \\
          &       & \cblue{Time(s) } & \cblue{\textbf{3} } & \cblue{1923}  & \cblue{25}    & \cblue{1246}  \\
          & \multirow{3}[0]{*}{30} & RMSE  & 0.0646  & 0.0485  & 0.0322  & \textbf{0.0146 } \\
          &       & RE    & 0.0278  & 0.0252  & 0.0178  & \textbf{0.0170 } \\
          &       &\cblue{ Time(s)}  & \cblue{\textbf{3 }} & \cblue{1930}  & \cblue{48}    & \cblue{1330}  \\
          & \multirow{3}[1]{*}{40} & RMSE  & 0.0641  & 0.0479  & 0.0267  & \textbf{0.0141 } \\
          &       & RE    & 0.0227  & 0.0194  & 0.0071  & \textbf{0.0056 } \\
          &       & \cblue{Time(s)}  & \cblue{\textbf{3 }} & \cblue{1911}  & \cblue{66}    & \cblue{706}  \\
    \midrule
    \multirow{12}[2]{*}{image2} & \multirow{3}[1]{*}{15} & RMSE  & 0.1050  & 0.0881  & 0.0755  & \textbf{0.0453 } \\
          &       & RE    & 0.1081  & 0.1066  & 0.1015  & \textbf{0.1007 } \\
          &       & \cblue{Time(s)}  & \cblue{\textbf{3 }} & \cblue{1630}  & \cblue{29}    & \cblue{833}  \\
          & \multirow{3}[0]{*}{20} & RMSE  & 0.1011  & 0.0815  & 0.0627  & \textbf{0.0305 } \\
          &       & RE    & 0.0686  & 0.0663  & 0.0579  & \textbf{0.0568 } \\
          &       & \cblue{Time(s)}  & \cblue{\textbf{3 }} & \cblue{1779}  & \cblue{40}    & \cblue{906}  \\
          & \multirow{3}[0]{*}{30} & RMSE  & 0.0993  & 0.0788  & 0.0511  & \textbf{0.0233 } \\
          &       & RE    & 0.0426  & 0.0386  & 0.0210  & \textbf{0.0185 } \\
          &       & \cblue{Time(s)}  & \cblue{\textbf{3 }} & \cblue{1841}  & \cblue{62}    & \cblue{585}  \\
          & \multirow{3}[1]{*}{40} & RMSE  & 0.0991  & 0.0785  & 0.0489  & \textbf{0.0224 } \\
          &       & RE    & 0.0390  & 0.0347  & 0.0121  & \textbf{0.0075 } \\
          &       & \cblue{Time(s)}  & \cblue{\textbf{3 }} & \cblue{1817}  & \cblue{66}    & \cblue{1349}  \\
    \midrule
    \multirow{12}[2]{*}{image3} & \multirow{3}[1]{*}{15} & RMSE  & 0.0910  & 0.0769  & 0.0689  & \textbf{0.0444 } \\
          &       & RE    & 0.1031  & 0.1020  & 0.0989  & \textbf{0.0980 } \\
          &       & \cblue{Time(s)}  & \cblue{\textbf{3 }} & \cblue{1735}  & \cblue{24}    & \cblue{880}  \\
          & \multirow{3}[0]{*}{20} & RMSE  & 0.0854  & 0.0685  & 0.0550  & \textbf{0.0286 } \\
          &       & RE    & 0.0634  & 0.0616  & 0.0560  & \textbf{0.0552 } \\
          &       & \cblue{Time(s)}  & \cblue{\textbf{3 }} & \cblue{1895}  & \cblue{34}    & \cblue{1082}  \\
          & \multirow{3}[0]{*}{30} & RMSE  & 0.0832  & 0.0653  & 0.0417  & \textbf{0.0198 } \\
          &       & RE    & 0.0356  & 0.0324  & 0.0194  & \textbf{0.0178 } \\
          &       & \cblue{Time(s)}  & \cblue{\textbf{3 }} & \cblue{1878}  & \cblue{58}    & \cblue{677}  \\
          & \multirow{3}[1]{*}{40} & RMSE  & 0.0830  & 0.0651  & 0.0387  & \textbf{0.0186 } \\
          &       & RE    & 0.0315  & 0.0279  & 0.0099  & \textbf{0.0067 } \\
          &       & \cblue{Time(s)}  & \cblue{\textbf{3 }} & \cblue{1893}  & \cblue{66}    & \cblue{667}  \\
    \bottomrule
      \label{tab:SNR1}%
    \end{tabular}}%

\end{table}%

%Table \Rmnum{1} summarizes the experimental results of all \cblue{\sout{the evaluating} unmixing} methods, under various noise levels  namely, \cblue{SNRs = \{15, 20, 30, 40\} dB}. 
% Table generated by Excel2LaTeX from sheet 'Sheet4'
\begin{table}[htbp]
  \centering
  \caption{\cblue{The Evaluate Results of the Proposed Method and the MV-NTF}}
  \cblue{
    \begin{tabular}{ccccc}
    \toprule
    Scenario & SNR   & Metric & MV-NTF & LR-NTF-MVNTF (ours) \\
    \midrule
    \multirow{12}[2]{*}{image1} & \multirow{3}[1]{*}{15} & RMSE  & 0.0831  & \textbf{0.0816 } \\
          &       & RE    & 0.0984  & \textbf{0.0951 } \\
          &       & Time(s)  & 1299  & \textbf{489 } \\
          & \multirow{3}[0]{*}{20} & RMSE  & 0.0693  & \textbf{0.0669 } \\
          &       & RE    & 0.0575  & \textbf{0.0534 } \\
          &       & Time(s)  & 1301  & \textbf{506 } \\
          & \multirow{3}[0]{*}{30} & RMSE  & 0.0713  & \textbf{0.0643 } \\
          &       & RE    & 0.0262  & \textbf{0.0170 } \\
          &       & Time(s)  & 1300  & \textbf{708 } \\
          & \multirow{3}[1]{*}{40} & RMSE  & 0.0719  & \textbf{0.0649 } \\
          &       & RE    & 0.0198  & \textbf{0.0057 } \\
          &       & Time(s)  & 1293  & \textbf{875 } \\
    \midrule
    \multirow{12}[2]{*}{image2} & \multirow{3}[1]{*}{15} & RMSE  & \textbf{0.0752 } & 0.0791  \\
          &       & RE    & 0.1029  & \textbf{0.1001 } \\
          &       & Time(s)  & 1297  & \textbf{494 } \\
          & \multirow{3}[0]{*}{20} & RMSE  & 0.0713  & \textbf{0.0694 } \\
          &       & RE    & 0.0587  & \textbf{0.0563 } \\
          &       & Time(s)  & 1298  & \textbf{609 } \\
          & \multirow{3}[0]{*}{30} & RMSE  & 0.0662  & \textbf{0.0589 } \\
          &       & RE    & 0.0221  & \textbf{0.0179 } \\
          &       & Time(s)  & 1304  & \textbf{680 } \\
          & \multirow{3}[1]{*}{40} & RMSE  & 0.0766  & \textbf{0.0707 } \\
          &       & RE    & 0.0155  & \textbf{0.0060 } \\
          &       & Time(s)  & 1304  & \textbf{652 } \\
    \midrule
    \multirow{12}[2]{*}{image3} & \multirow{3}[1]{*}{15} & RMSE  & \textbf{0.0800 } & 0.0847  \\
          &       & RE    & 0.1013  & \textbf{0.0975 } \\
          &       & Time(s)  & 1299  & \textbf{465 } \\
          & \multirow{3}[0]{*}{20} & RMSE  & 0.0683  & \textbf{0.0657 } \\
          &       & RE    & 0.0592  & \textbf{0.0548 } \\
          &       & Time(s)  & 1297  & \textbf{536 } \\
          & \multirow{3}[0]{*}{30} & RMSE  & 0.0722  & \textbf{0.0650 } \\
          &       & RE    & 0.0277  & \textbf{0.0174 } \\
          &       & Time(s)  & 1297  & \textbf{755 } \\
          & \multirow{3}[1]{*}{40} & RMSE  & 0.0676  & \textbf{0.0597 } \\
          &       & RE    & 0.0217  & \textbf{0.0058 } \\
          &       & Time(s)  & 1305  & \textbf{957 } \\
    \bottomrule
    \end{tabular}%
    }
  \label{tab:MVNTF}%
\end{table}%

For all of the methods, the abundance matrix estimated by the FCLS algorithm \cite{Heinz2001} was used for the abundance initialization. Furthermore, as this was a supervised unmixing task, the endmember matrix  was supposed to be known beforehand. 
Since  the quality of the endmember extraction  affects the results of the abundance inversion, for all the scenes shown in Table \ref{tab:SNR1},  the true endmember information was used. The interaction endmember signals were obtained from the corresponding Hadamard product of two true endmembers. 
The parameters of LR-NTF were set as follows: $\lambda _1=0.1$,  $\lambda _2=0.07$, and    $\mu=8\times 10^{-3}$.  Meanwhile, the maximum number of program iterations was set to 1000.
We remark that theoretically, under the same constraints, the optimal solution of  LR-NTF (whose objective function is a combination of the reconstruction error and two regularization terms)  is expected to yield a higher reconstruction error than the solution of the Semi-NMF (whose objective function is the single reconstruction error). However, the exact optimal solutions for both above two problems are hard to obtain in practice. Semi-NMF and LR-NTF obtain approximated optimal solutions. And the feasibility of these solutions is not guaranteed. So that it is possible for the solution of LR-NTF to be better in the aspect of the reconstruction error.

\begin{figure}[htbp]
\centering
\includegraphics[scale=0.33]{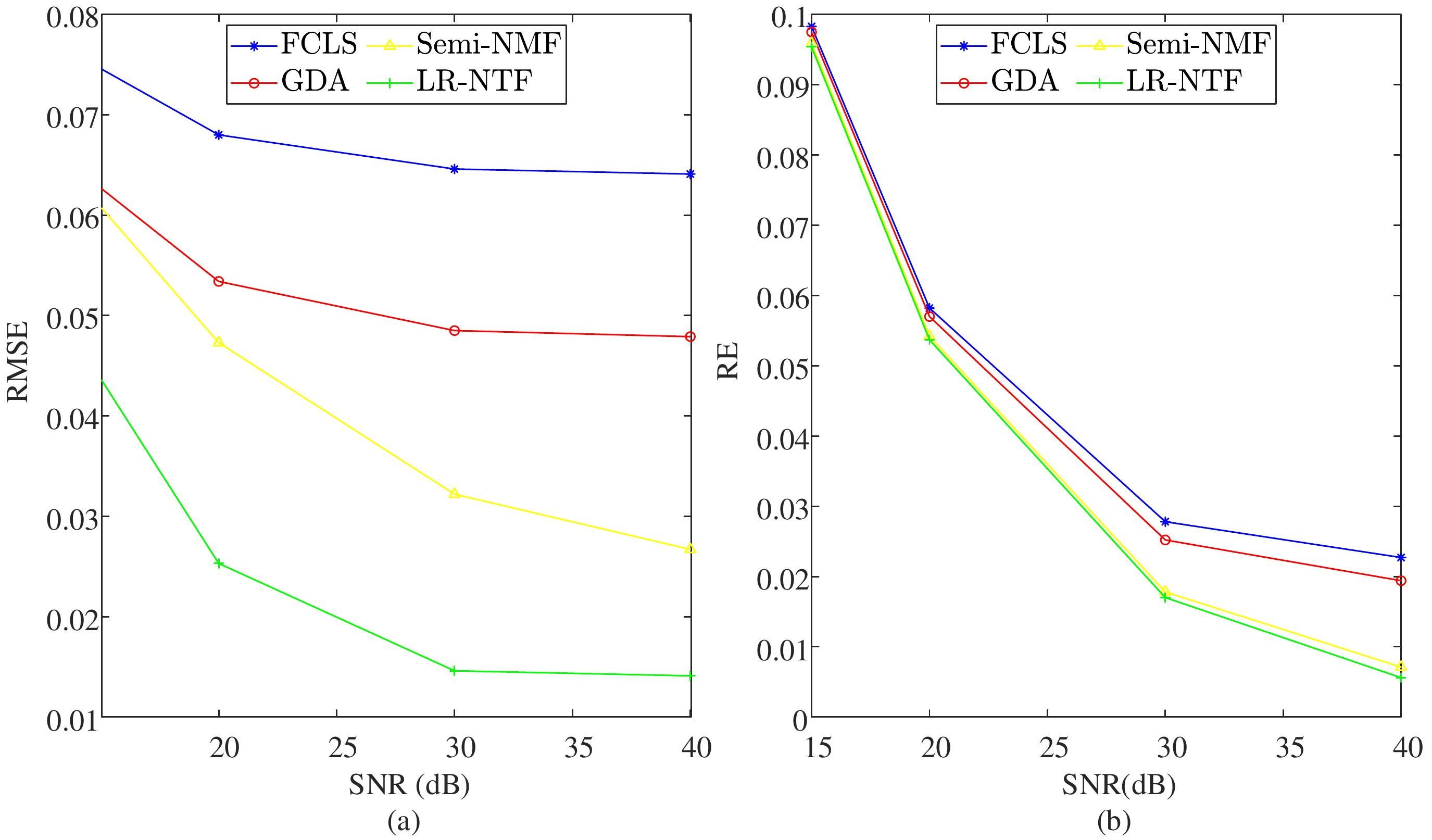}
\caption{Unmixing performance in terms of RMSE (a) and RE (b) in the simulated image1 with different SNRs}
\label{fig:SNR}
\end{figure}

\cblue{It can be seen} from Fig. \ref{fig:SNR} that, for the different levels of noise in image1, the proposed method  uniformly yields the best performance in terms of RMSE and RE. Of the three nonlinear unmixing algorithms, the GDA produces the worst results since it carries out the unmixing pixel by pixel without considering spatial   correlation or sparsity of abundances. The semi-NMF and the proposed LR-NTF  use  nonnegative  matrix  factorization and tensor factorization, respectively,  to  process  a  whole image. These two kinds of factorization  exploit the spatial correlation of HSIs. Furthermore, the proposed method takes advantage of the low rank of abundance maps, which boosts its unmixing performance. 

To evaluate the robustness of the proposed algorithm to the different types of spectral mixing, we generated three images:  image1, which was based on the  GBM, image2 based on the PPNM, and image3, for which half the pixels were based on the  GBM and the other half were based on the PPNM.
It can be seen from Table \ref{tab:SNR1} that, although it was  based on the GBM model, the proposed LR-NTF performed best not only when applied to image1, but also for image2 and image3.
 
\begin{figure}[htbp]
\centering
\includegraphics[scale=0.31]{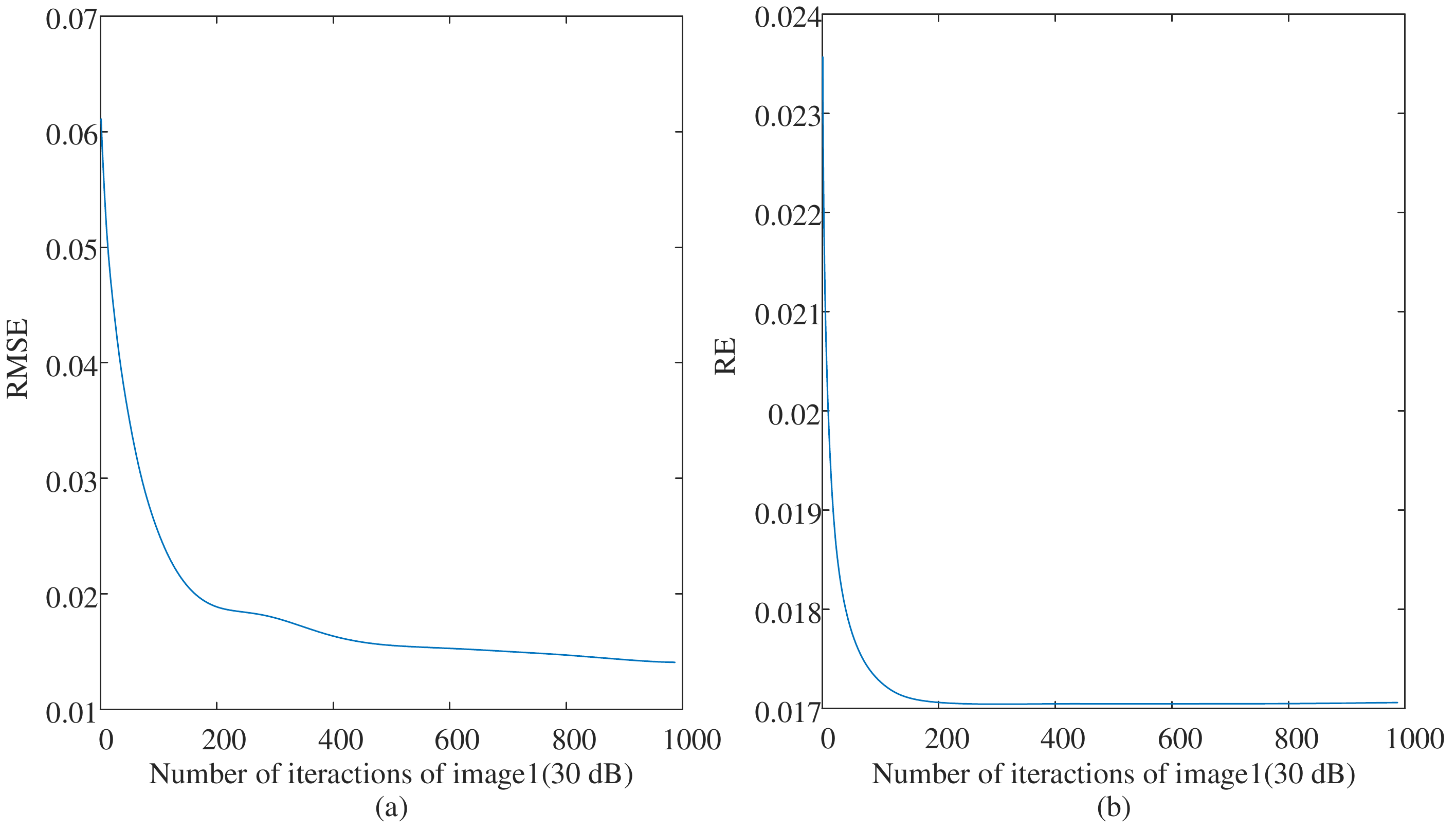}
\caption{Iterations of RMSE (a) and RE (b) in LR-NTF with the simulated image1}
\label{fig:image1}
\end{figure}
\begin{figure}[htbp]
\centering
\includegraphics[scale=0.31]{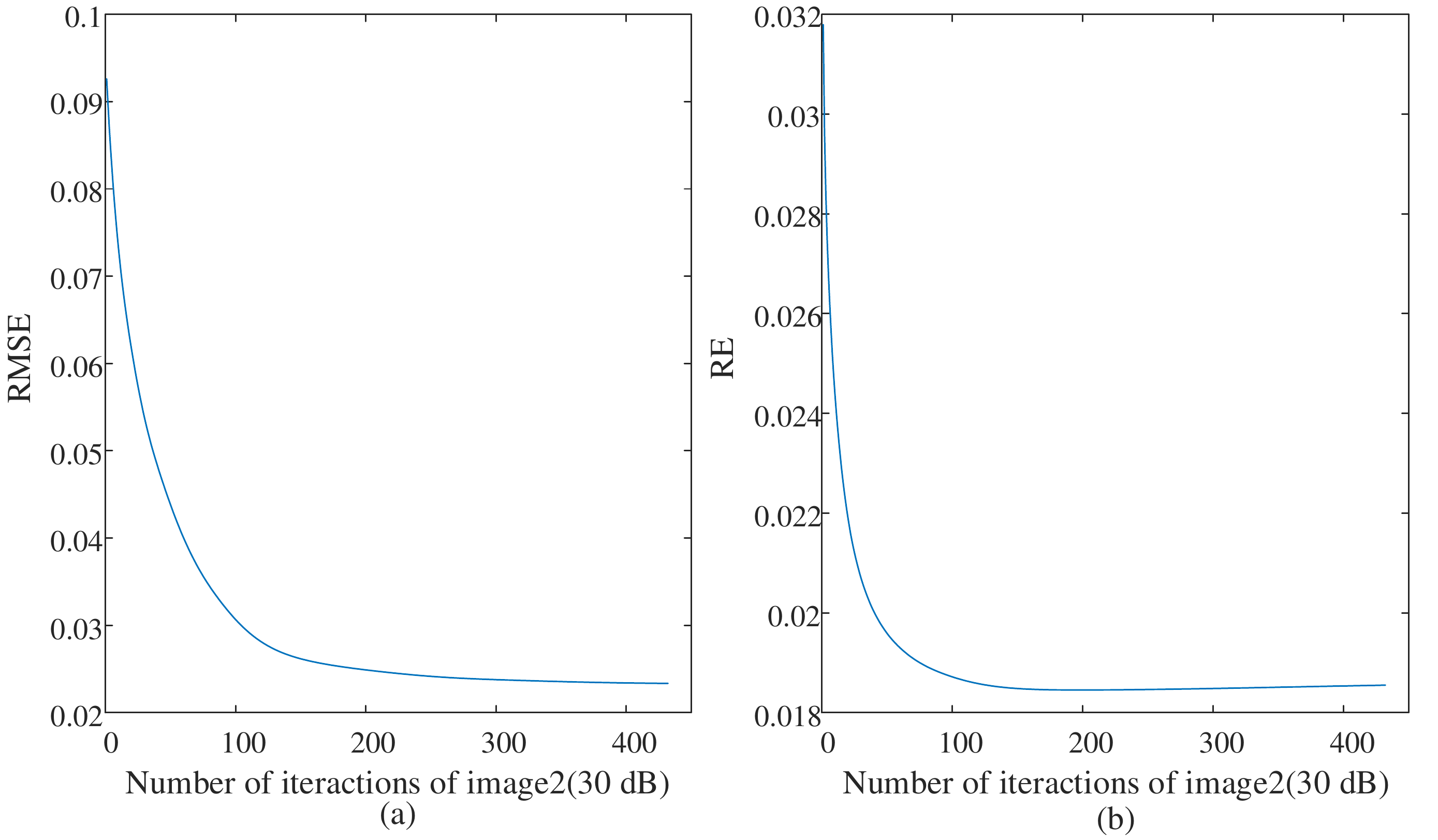}
\caption{Iterations of RMSE (a) and RE (b) in LR-NTF with the simulated image2}
\label{fig:image2}
\end{figure}
\begin{figure}[htbp]
\centering
\includegraphics[scale=0.31]{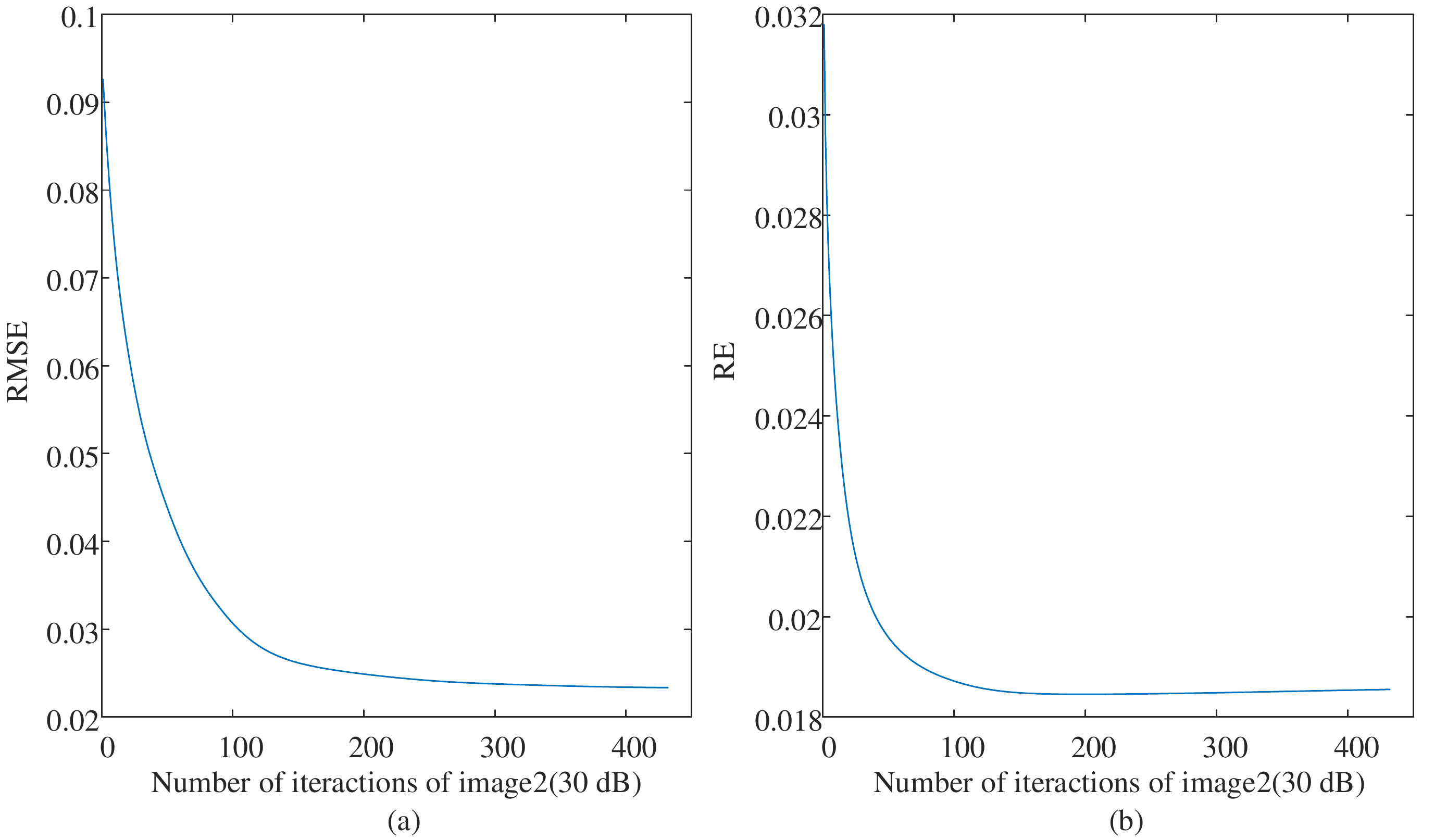}
\caption{Iterations of RMSE (a) and RE (b) in LR-NTF with the simulated image3}
\label{fig:image3}
\end{figure}

To evaluate the impact of low-rank representations of interaction abundance maps in nonlinear mixing images, we conducted a comparison between our  method and the MV-NTF method \cite{Qian2017}, which unmixes image based on LMM model by using low-rank representations of abundance maps. MV-NTF method estimates abundances and endmembers jointly, whereas the proposed LR-NTF only estimates abundances. For a fair comparison, a modified version of LR-NTF method, called 'LR-NTF-MVNTF', was  implemented using endmembers estimated by MV-NTF, instead of being estimated by VCA. Results in Table \ref{tab:MVNTF} show that our 
method, LR-NTF-MVNTF, \cblue{ can obtain the best results in most cases with
gains increasing as the SNR increases.}

Fig. \ref{fig:image1}, Fig. \ref{fig:image2},and Fig. \ref{fig:image3} show the iterations of RMSE and RE with the image1 (30 dB), image2 (30 dB), image3 (30 dB),  respectively\cite{Yokoya2014}. It is shown that the proposed algorithm is an excellent algorithm with good convergence.

\cblue{As can be seen in Table \ref{tab:SNR1}, the FCLS obtains the least run time, and the GDA obtains the longest running time.  Although the proposed method is based on the NTF, it's also faster than GDA and MV-NTF.}

\section{ Experiments with Real Dataset}
%In this section, the proposed tensor-based nonlinear unmixing algorithm is tested in four real images. Two of the images are captured by the Airborne Visible Infrared Imaging Spectrometer (AVIRIS) sensor, and the others are collected with the Hyperspectral Digital Imagery Collection Experiment (HYDICE) sensor. Due to the lack of the ground truth of abundances, RE in (\ref{eq:RE}) is used to measure the performance of the proposed method and the state-of-the-art methods.
In this section, We describe tests where the proposed tensor-based
nonlinear unmixing algorithm was applied to four real images. Two of the images were acquired by the Airborne Visible Infrared Imaging Spectrometer (AVIRIS), and the others were collected by the Hyperspectral Digital Imagery Collection Experiment (HYDICE) sensor. Due to the lack of the ground truth for the abundances, the RE in (\ref{eq:RE}) \cblue{and average of spectral angle mapper (aSAM) were} used to measure the performance of both the proposed method and the state-of-the-art methods. \cblue{The aSAM metric can qualify the average spectral angle mapping of the reconstructed  \textit{j}th spectral vector  $\hat{\textbf{y}}_{j}$ and observed   \textit{j}th spectral  vector  $\textbf{y}_{j}$:  }
\begin{equation}
\cblue{\text{aSAM} = \frac{1}{N}\sum\limits_{j = 1}^N \text{arccos} \left( {\frac{{\textbf{y}_{j}^\mathrm{T}\cdot\widehat{\textbf{y}}_{j}}}{{\left\| \textbf{y}_j \right\|\left\| \widehat{\textbf{y}}_{j} \right\|}}} \right).}
 \label{eq:SAM}
\end{equation}
Note that RE and aSAM  are   reference metrics, not   direct metrics   measuring the quality of estimated abundances.

\begin{figure*}[htbp]
\centering   
\subfigure[Cuprite]{
\label{fig:realdata11}
\includegraphics[width=0.296\textwidth]{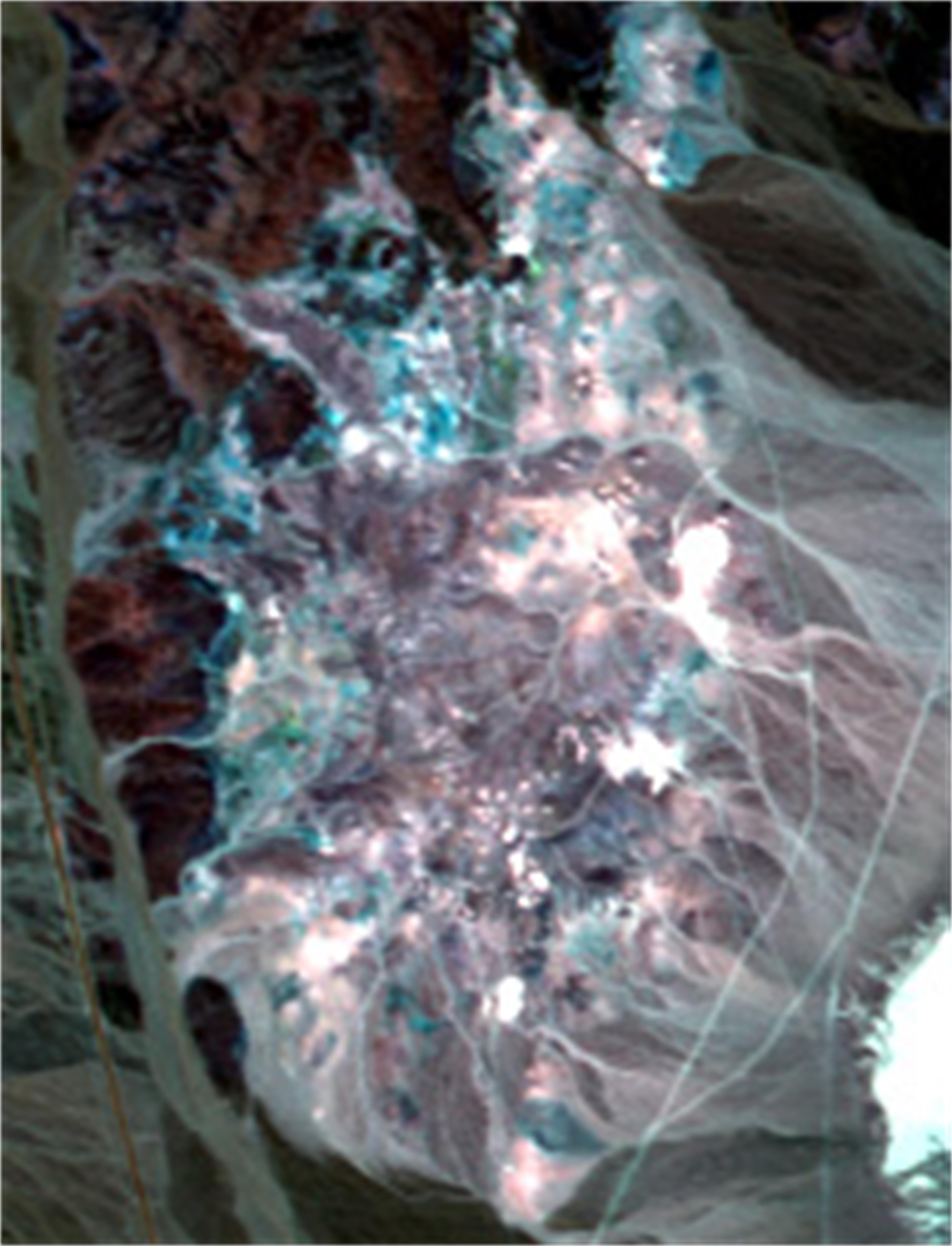}}
\subfigure[Reference Map of Cuprite]{
\label{fig:realdata12}
\includegraphics[width=0.4\textwidth]{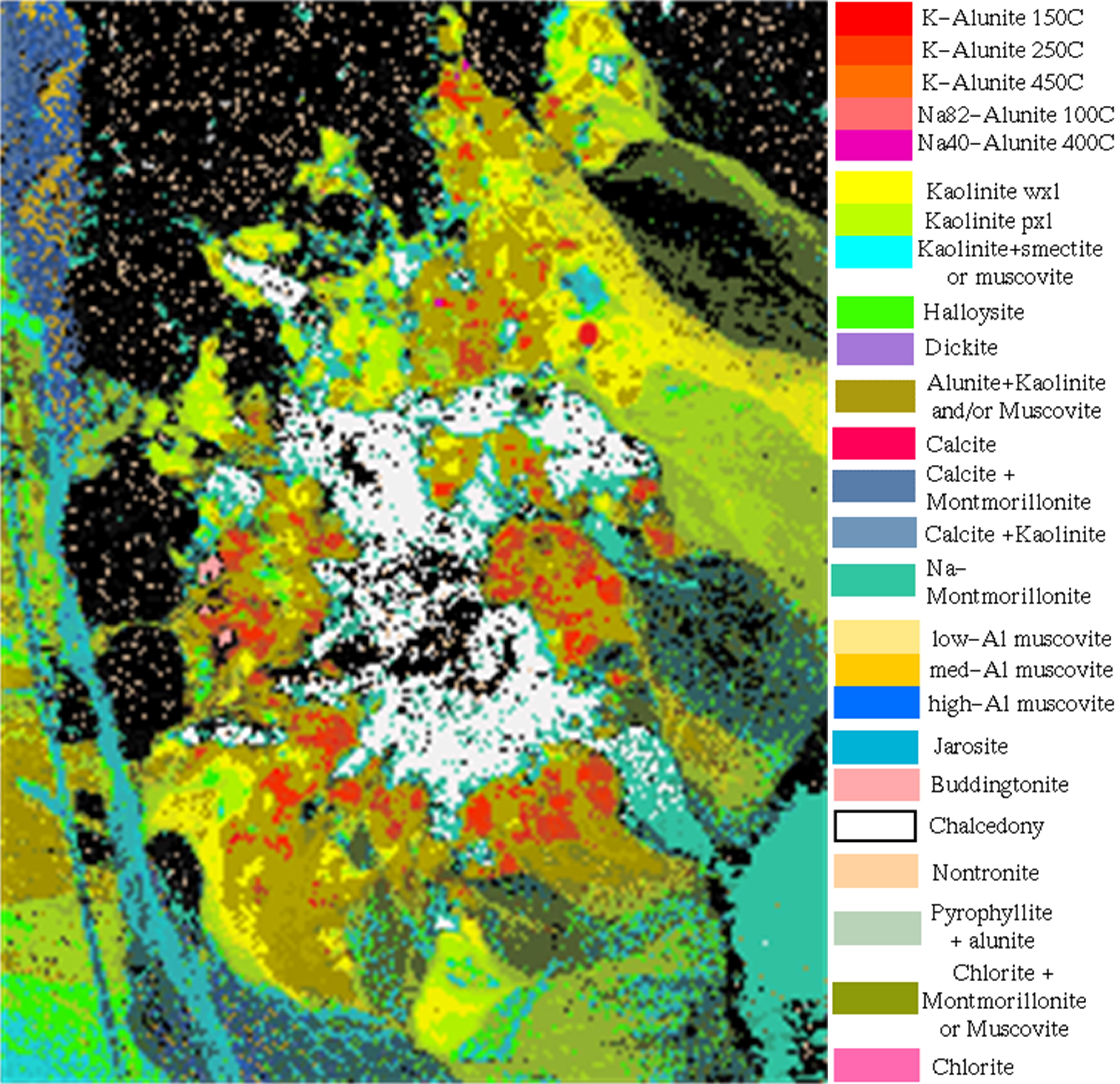}}
\subfigure[San Diego Airport]{
\label{fig:realdata2}
\includegraphics[width=0.3\textwidth]{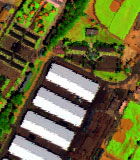}}
\subfigure[sub-DC1]{
\label{fig:realdata3}
\includegraphics[width=0.35\textwidth]{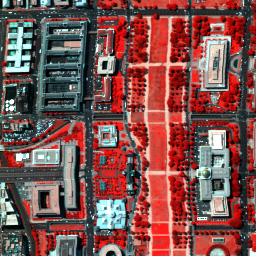}}
\subfigure[sub-DC2]{
\label{fig:realdata4}
\includegraphics[width=0.25\textwidth]{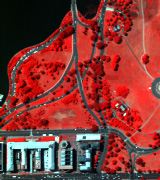}}
\caption{HSIs used in the experiments: (a) sub-image of Cuprite provided by USGS in 1995, (b) reference map of Cuprite, (c) sub-image of San Diego Airport data, (d) sub-image1 of Washington Dc Mall, (e) sub-image2 of Washington Dc Mall}
\label{fig:realdata}
\end{figure*} 
\subsection{Cuprite}
The first of the real datasets was the widely used hyperspectral image for testing unmixing methods\cite{Qian2011,Wei2017,Wang2017,Zhuang2019}, which was acquired by the AVIRIS sensor over the Cuprite mining region, NV, USA. The image has 224 spectral bands in the range  $0.4\sim2.5\mu$m. After removing the bands with a low SNR and the water absorption bands, 188 bands remained. A sub-image with $250\times191$ pixels was chosen for use in our experiments: refer to Fig. \ref{fig:realdata}(a) for details. \cblue{The spatial distribution of minerals can be inferred from a reference mineral map (shown in Fig. \ref{fig:realdata}(b)) created by
a classifier based on domain-specific knowledge.}

The sub-image that was used has been extensively studied in \cite{Qian2011,Wei2017,Wang2017,Zhuang2019}. We set the number of endmembers, R, to 10, the same as in \cite{Qian2011}, and \cite{Li2016a}. Meanwhile, VCA\cite{Nascimento2005} was used to extract the endmember signatures, which were Andradite, Sphene, Muscovite, Montmorillonite CM20, Kaolinite, Alunite, Nontronite, Montmorillonite Na, Buddingtonite, and Chalcedony.  

\begin{figure}[htbp]
\centering
\includegraphics[scale=0.33]{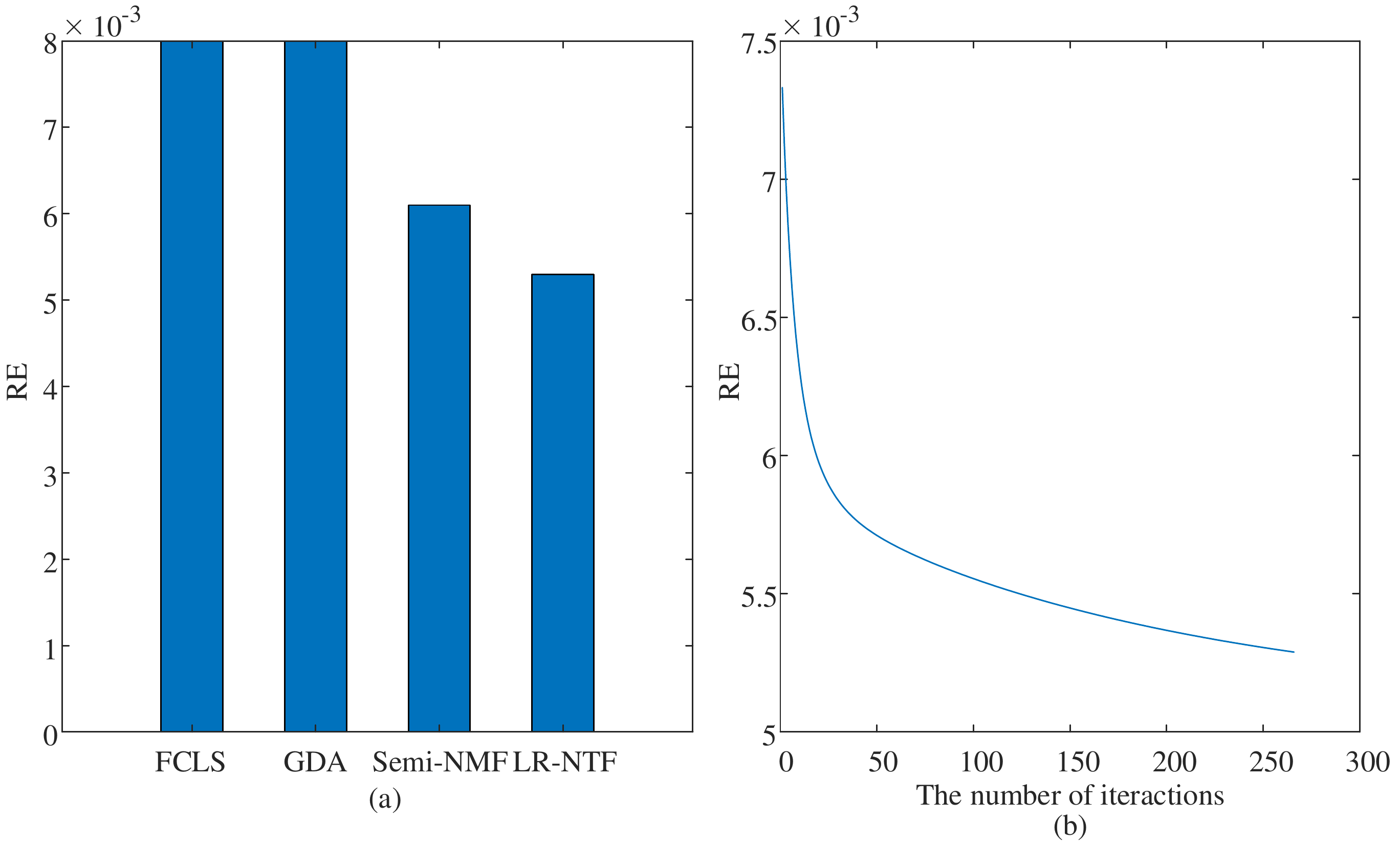}
\caption{Image reconstruction errors (REs) and the convergence of LR-NTF in Cuprite data}
\label{fig:re-cup}
\end{figure}

The  ASC regularization term in the semi-NMF and \cblue{MV-NTF were set  to 0.1 and 0.6, respectively.} For the proposed LR-NTF method, the low-rank abundance parameter  and the nonlinear interaction regularization terms were set to 0.1 and 0.07, respectively. Furthermore, the penalty parameter $\mu=1\times10^{-4}$, and the abundance maps were initialized using the FCLS algorithm.  \cblue{The tolerance level, at which the iterations were stopped, was set to $1\times 10^{-6}$ for \cgreen{GDA, semi-NMF, and LR-NTF}. }
The number of iterations of the proposed  LR-NTF was 260 for the Cuprite image (see   Fig. \ref{fig:re-cup}(b)).

% Table generated by Excel2LaTeX from sheet 'Sheet6'
\begin{table*}[htbp]
  \centering
  \caption{\cblue{Evaluation Results in Cuprite with RE, aSAM and time cost (s).}}
  \cblue{
    \begin{tabular}{cccccc|cc}
    \toprule
    \multirow{2}[4]{*}{Scenario} & \multirow{2}[4]{*}{Metric} & FCLS  & GDA   & Semi-NMF & LR-NTF (ours) & MV-NTF & LR-NTF-MVNTF (ours) \\
\cmidrule{3-8}          &       & \multicolumn{4}{c|}{Using endmembers extracted by VCA} & \multicolumn{2}{c}{Using endmembers extracted by MVNTF} \\
    \midrule
    \multirow{3}[2]{*}{Cuprite} & RE    & 0.0080  & 0.0080  & 0.0061  & \textbf{0.0053 } & 0.0104  & \textbf{0.0038 } \\
          & aSAM  & 0.0191  & 0.0191  & 0.0165  & \textbf{0.0145 } & 0.0134  & \textbf{0.0107 } \\
          & Time  & \textbf{35 } & 176   & 93    & 397   & 4643  & 187 \\
    \bottomrule
    \end{tabular}%
    }
  \label{tab:addcup}%
\end{table*}%

\begin{figure*}[htbp]
\centering
\includegraphics[scale=0.60]{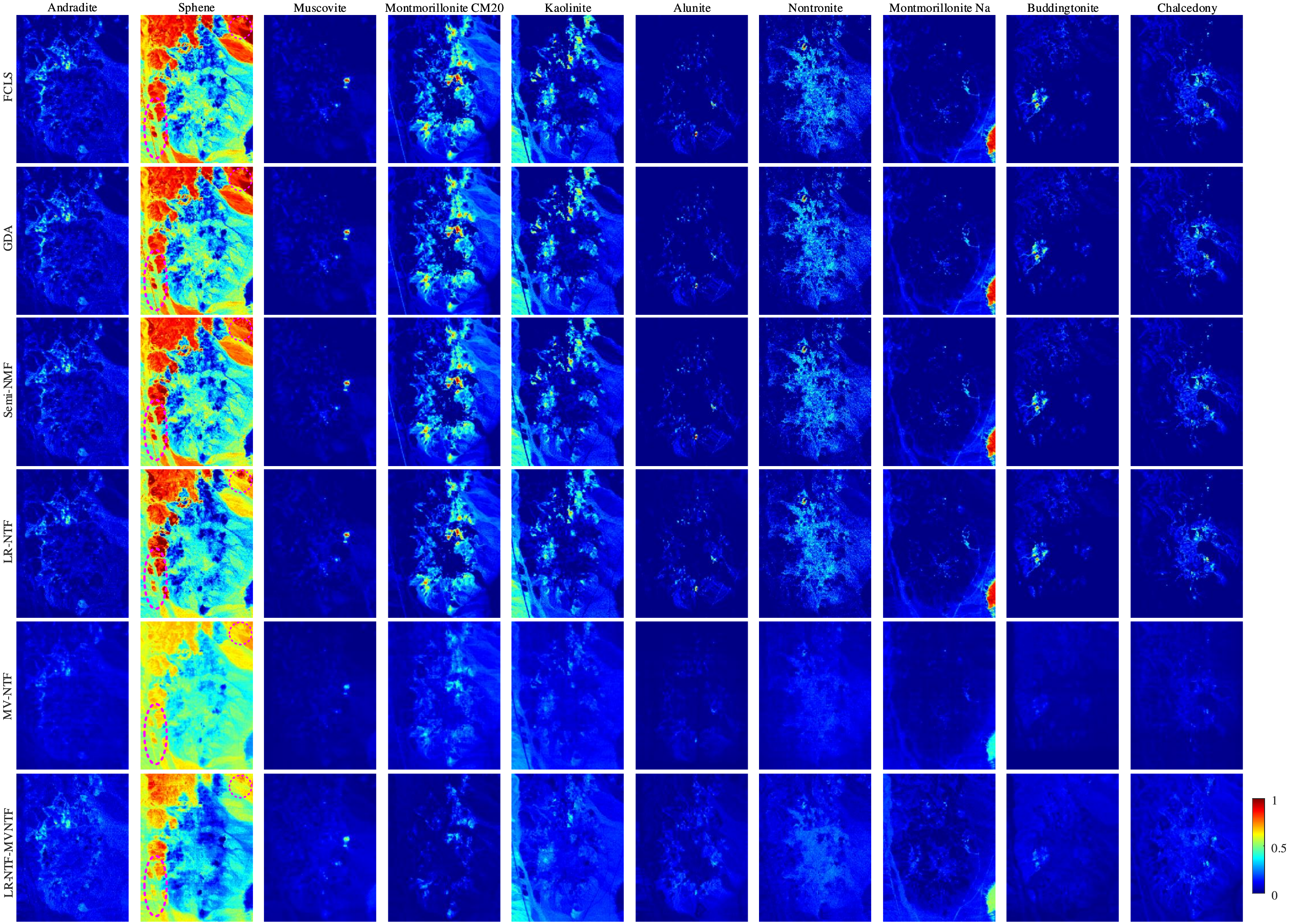}
\caption{Estimated abundance maps of the Cuprite image}
\label{fig:abu-cup}
\end{figure*}

\begin{figure}[htbp]
\centering
\includegraphics[scale=0.28]{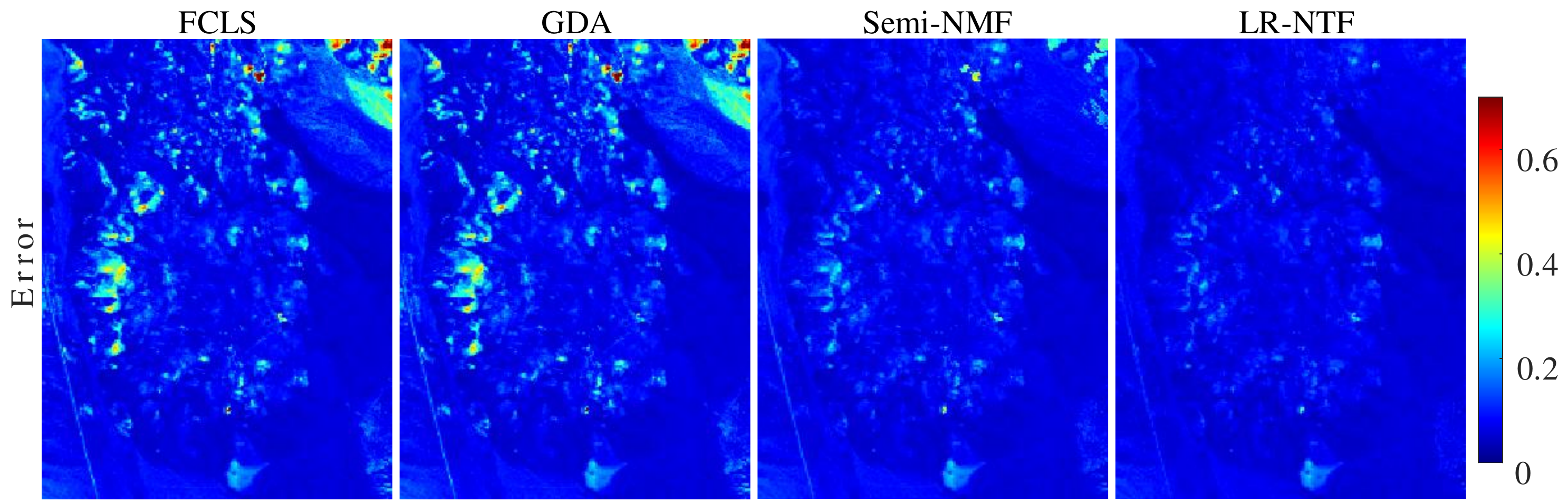}
\caption{RE distribution maps comparison between the proposed algorithm and state-of-the-art algorithms on the Cuprite}
\label{fig:err-cup}
\end{figure}

\cblue{Table \ref{tab:addcup} shows the RE, aSAM, and time cost  for all the algorithms.} It can be seen that, as it considers the spatial and low-rank information in the HSIs simultaneously, the error produced by the proposed LR-NTF is the lowest. \cblue{Meanwhile, the proposed LR-NTF using endmembers extrated by MV-NTF obtains   lower RE and aSAM than  the LR-NTF using endmembers extrated by VCA.}
Fig. \ref{fig:abu-cup} and Fig. \ref{fig:err-cup} show  the estimated abundance maps and distributions of the REs, respectively. \cblue{The significant differences of our results compared to other methods in Fig. \ref{fig:abu-cup} are marked in carmine circles.} 
The bright areas in Fig. \ref{fig:err-cup}  indicate  large errors in the reconstructed images. The errors for the FCLS are the worst because this method only considers linear mixtures of the minerals. The semi-NMF performs better than GDA because the GDA is a pixel-based algorithm that does not take any spatial information into consideration. As can be seen from Fig. \ref{fig:err-cup}, the proposed algorithm performs significantly better than the semi-NMF as it takes into account  spatial information from the region of interest by using the tensor-based form and low-rank distribution of materials.

\subsection{San Diego Airport}
The second real dataset used in the experiment was an airborne hyperspectral image of San Diego Airport that was acquired by the AVIRIS sensor. This image has been widely used in hyperspectral target detectionn\cite{Taghipour2017,Vafadar2018}. The image contains $400\times400$ pixels and 224 spectral bands. As in\cite{zhu2017spectral}, a sub-image with $160\times140$ pixels was chosen  for our experiments,  and is shown in Fig. \ref{fig:realdata2}. In order to avoid the errors caused by water vapor absorption and low SNRs, bands 1-6, 33-35, 97, 107-113,  153-166, and 221-224 were removed. The sub-image included the following five main materials: Roof, Grass, Ground and Road, Tree, and Other (see \cite{zhu2017spectral} for more details).  We, therefore, set $R=5$ in the experiments; the VCA was used to extract the endmembers.

\begin{figure}[htbp]
\centering
\includegraphics[scale=0.31]{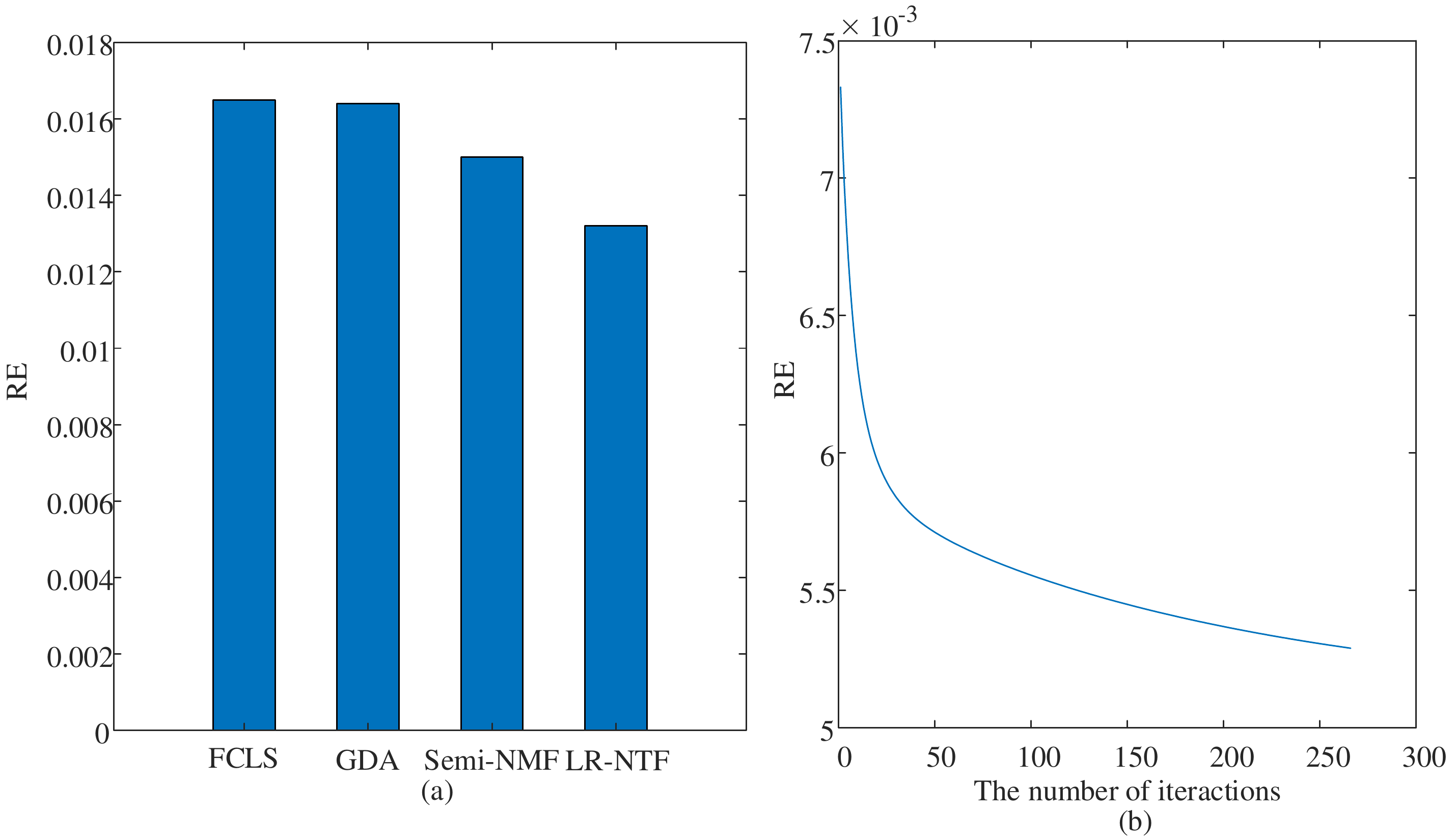}
\caption{RE of four unmixing methods in (a) and iteration of RE in the LR-NTF in (b) with the San Diego Airport data}
\label{fig:re-san}
\end{figure}

The optimal parameter values were used in the comparison experiments. The ANC balance parameters in the semi-NMF \cblue{ and MV-NTF were set to 0.1 and 0.6, respectively}. For the proposed algorithm, the low-rank abundance regularization parameter was set to 0.1, the low-rank nonlinear interaction regularization parameter was set to 0.07, and the penalty parameter to $1\times10^{-4}$. Furthermore, the FCLS was used to initialize the abundance map, and the maximum number of iterations was set to 1000. The tolerance used for stopping the iterations of a solver was set to $1\times 10^{-6}$ for \cgreen{GDA, semi-NMF, and LR-NTF}. Fig. \ref{fig:re-san}(b) shows a plot of RE against the number of iterations for the San Diego Airport data. \cblue{Table \ref{tab:addsan} shows the RE and aSAM  for all the algorithms.} It can be seen that, as it considers the spatial and low-rank information in the HSIs simultaneously, the error produced by the proposed LR-NTF is the lowest. A comparison between results of LR-NTF (using endmembers estimated by MV-NTF)  and LR-NTF-MVNTF using endmembers estimated by VCA)  indicates that the performance of the proposed LR-NTF can be improved by using a  higher quality of endmembers.

% Table generated by Excel2LaTeX from sheet 'Sheet6'
\begin{table*}[htbp]
  \centering
  \caption{\cblue{Evaluation Results in San Diego Airport with RE, aSAM and time cost (s).}}
  \cblue{
    \begin{tabular}{cccccc|cc}
    \toprule
    \multirow{2}[4]{*}{Scenario} & \multirow{2}[4]{*}{Metric} & FCLS  & GDA   & Semi-NMF & LR-NTF (ours) & MV-NTF & LR-NTF-MVNTF (ours) \\
\cmidrule{3-8}          &       & \multicolumn{4}{c|}{Using endmembers extracted by VCA} & \multicolumn{2}{c}{Using endmembers extracted by MVNTF} \\
    \midrule
    \multirow{3}[2]{*}{San Diego Airport} & RE    & 0.0165  & 0.0164  & 0.0150  & \textbf{0.0132 } & 0.0184  & \textbf{0.0110 } \\
          & aSAM  & 0.0596  & 0.0594  & 0.0542  & \textbf{0.0455 } & 0.0588  & \textbf{0.0412 } \\
          & Time  & \textbf{4 } & 193   & 50    & 62    & 4946  & 188 \\
    \bottomrule
    \end{tabular}%
    }
  \label{tab:addsan}%
\end{table*}%

\begin{figure}[htbp]
\centering
\includegraphics[scale=0.90]{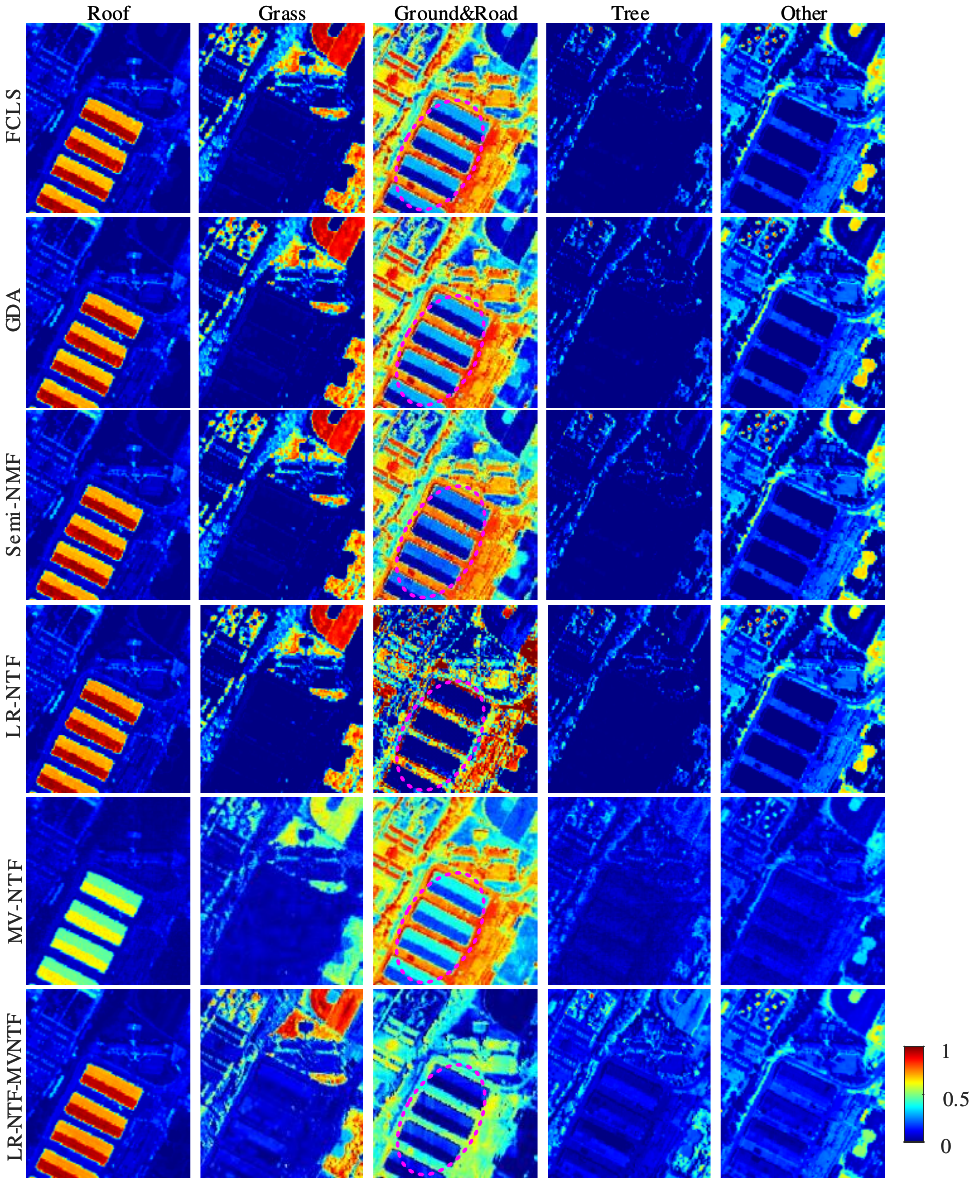}
\caption{Estimated abundance maps comparison between the proposed algorithm and state-of-the-art algorithms on the San Diego Airport}
\label{fig:abu-san}
\end{figure}

\begin{figure}[htbp]
\centering
\includegraphics[scale=0.29]{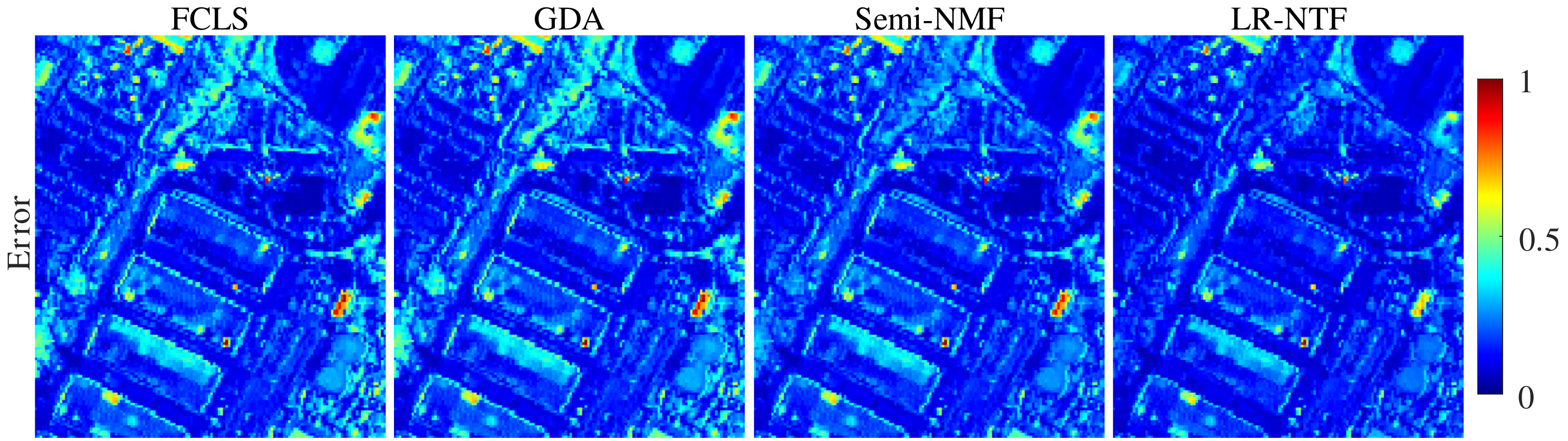}
\caption{ RE distribution maps comparison between the proposed algorithm and state-of-the-art algorithms on the San Diego Airport}
\label{fig:err-san}
\end{figure}

It can be seen from Fig. \ref{fig:abu-san} that the proposed algorithm successfully distinguishes between the Roof and the \cblue{Ground\&Road} classes \cblue{marked by carmine circles}. Fig. \ref{fig:re-san}(a) shows the RE values for all of the algorithms, and it can be seen that the proposed algorithm produces the best results. Further details are given in Fig. \ref{fig:err-san}, where the RE distribution is shown visually. The very bright areas in the image correspond to large REs: \cblue{it is clear from this image that the error map obtained using the proposed algorithm \cblue{shows} smaller error than the ones produced by the other algorithms.}

\subsection{Washington DC Mall}
The third real dataset used in the experiment was the Washington DC Mall scene, which was acquired by the HYDICE sensor over Washington DC, USA. The full image contains $1208\times307$ pixels, with 210 spectral bands ranging from 0.4 to 2.5 $\mu$m. The spatial resolution is 3 m. After removing the atmospheric absorption and low-SNR bands (bands 103–106, 138–148, and 207–210), 191 bands remained. This image contains seven main materials: Roof, Grass, Road, Trail, Water, Shadow, and Tree. Two sub-images were chosen from the original image to test the proposed algorithms. 

\begin{figure}[htbp]
\centering
\includegraphics[scale=0.3]{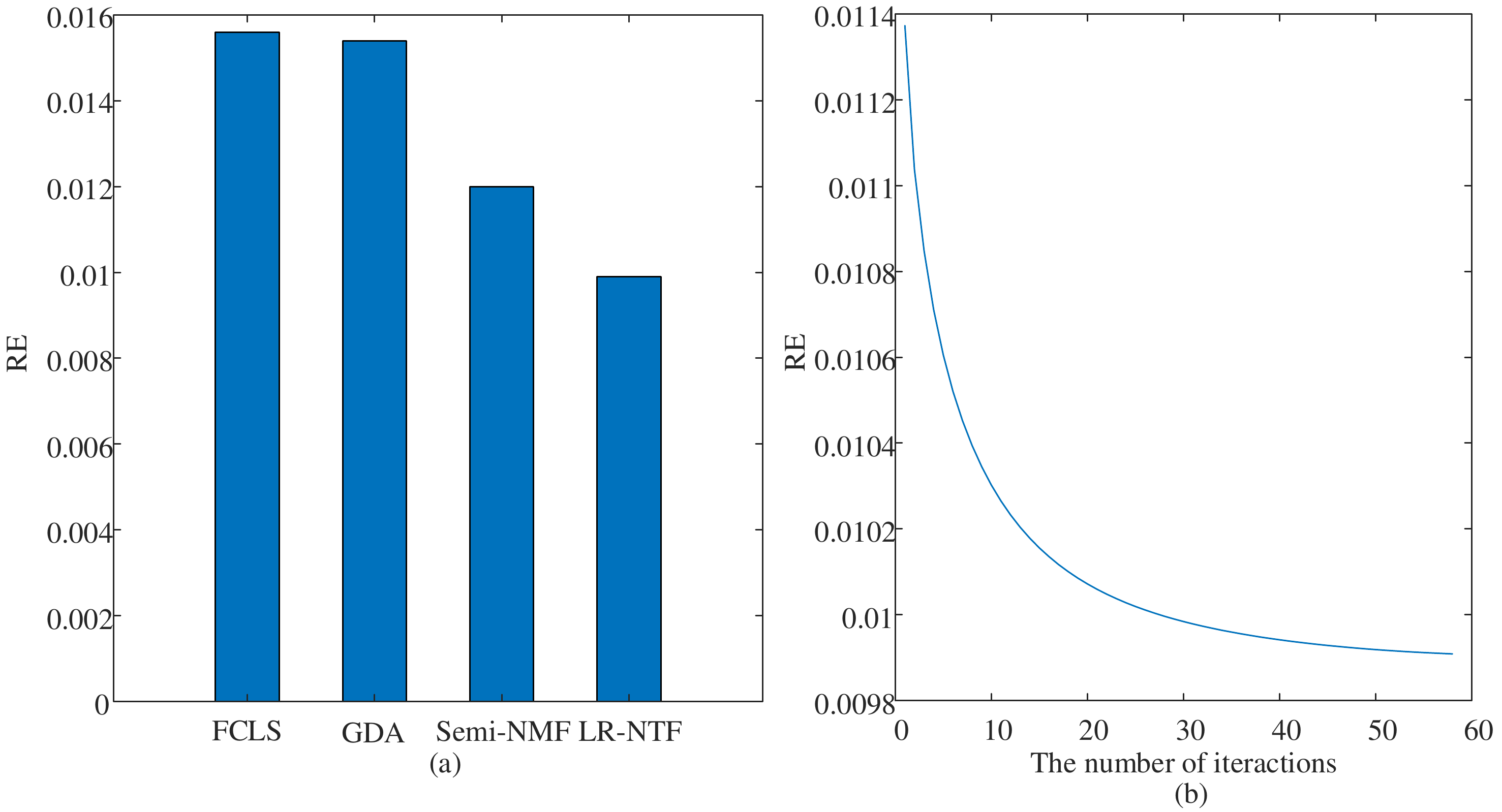}
\caption{RE of four unmixing methods in (a) and iteration of RE in the LR-NTF in (b) with the sub-image1 of  Washington DC Mall data}
\label{fig:re-wash1}
\end{figure}

\begin{figure*}[htbp]
\centering
\includegraphics[scale=0.87]{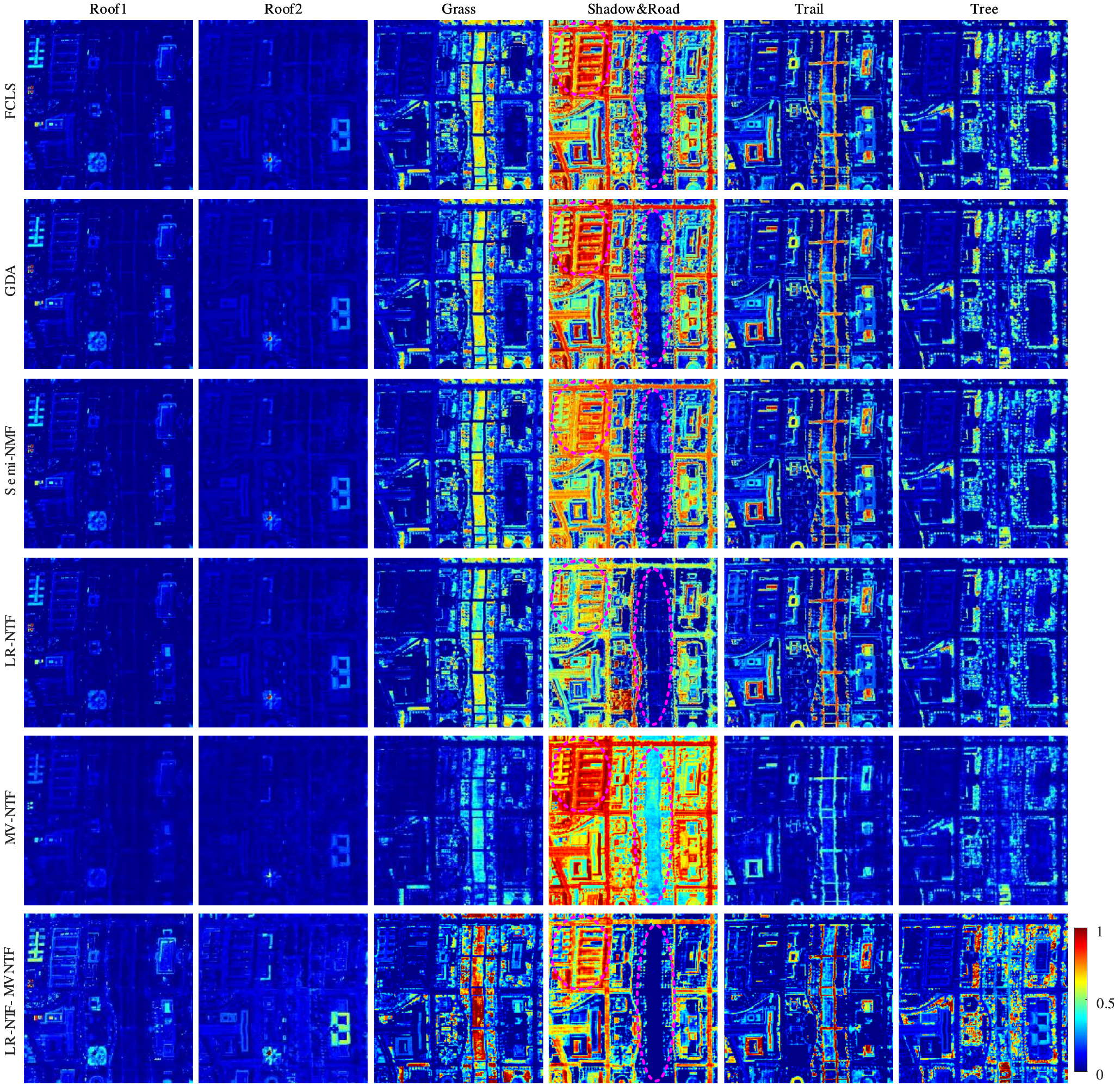}
\caption{Estimated abundance maps comparison between the proposed algorithm and state-of-the-art algorithms on sub-image1 of Washington DC Mall data}
\label{fig:abu-wash1}
\end{figure*}

\begin{figure}[htbp]
\centering
\includegraphics[scale=0.29]{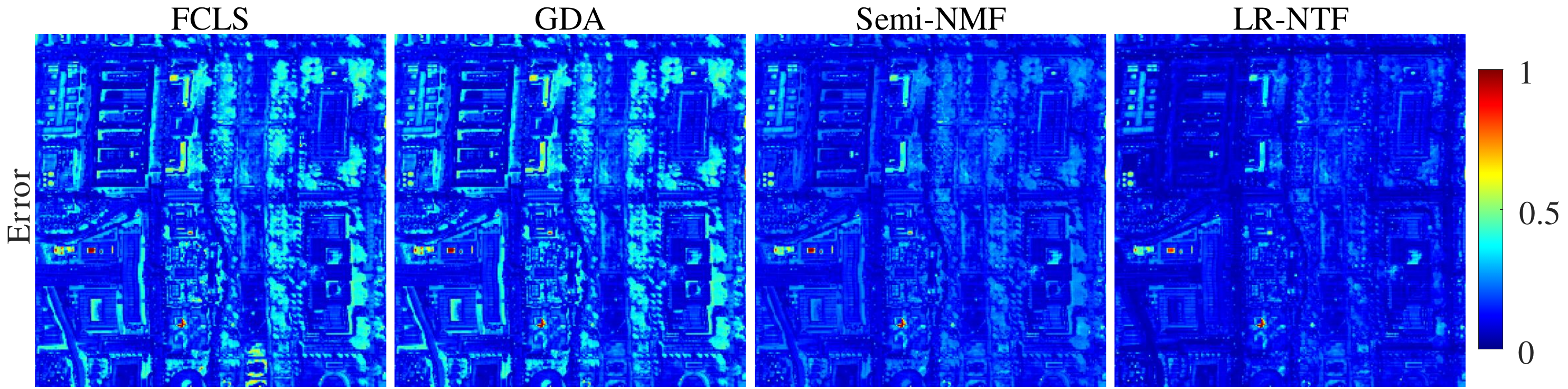}
\caption{  RE distribution maps comparison between the proposed algorithm and state-of-the-art algorithms on sub-image1 of Washington DC Mall data }
\label{fig:err-wash1}
\end{figure}

The first sub-image consisting of 256x256 pixels, called sub-DC1, was clipped from the Washington DC Mall data (see Fig. \ref{fig:realdata3}).  Hysime\cite{BioucasDias2008} and the VCA were used to estimate the number of endmembers and the endmember matrix, respectively.
The extracted endmembers were called Roof1, Roof2, Grass, Road, Tree, and Trail. The optimal parameter values for all of the algorithms were set as follows. The ASC parameter was set to 0.1 in the semi-NMF; for the proposed LR-NTF method, the low-rank regularization parameter was set to 0.1 in the abundance map and 0.07 in the nonlinear interaction map. The  LR-NTF penalty parameter was set to $1\times10^{-4}$, and FCLS was used to initialize the abundance maps. The tolerance level used to stop the iterations of a solver was set to $1\times 10^{-6}$ in \cgreen{GDA, semi-NMF, and LR-NTF}. Fig. \ref{fig:re-wash1} shows the RE for all the algorithms together with the RE plotted against the number of iterations of the proposed algorithm. Fig. \ref{fig:abu-wash1} depicts the estimated abundance maps for the proposed algorithm as well as the other algorithms. A detailed analysis shows that the FCLS produces poor estimates of the abundance maps since the LMM fails to model the complex information in the scene. Meanwhile, the semi-NMF performs better than the GDA as it takes into account some of the spatial information by using the hyperspectral matrix. Furthermore,  by considering the low-rank representation and the third-order tensor, the proposed method produces effective results. In order to study the differences in the error produced by all the algorithms, the RE distributions are shown visually in Fig. \ref{fig:err-wash1}. \cblue{The results in Table \ref{tab:addwash} illustrates that the proposed algorithm achieves a smaller error than other methods.} \cblue{After using the endmembers extrated by MV-NTF, the proposed LR-NTF obtains the  more reasonable abundance estimation results than that by VCA.}

% Table generated by Excel2LaTeX from sheet 'Sheet6'
\begin{table*}[htbp]
  \centering
  \caption{\cblue{Evaluation Results in Washinrton DC Mall with RE, aSAM and time cost (s).}}
  \cblue{
    \begin{tabular}{cccccc|cc}
    \toprule
    \multirow{2}[4]{*}{Scenario} & \multirow{2}[4]{*}{Metric} & FCLS  & GDA   & Semi-NMF & LR-NTF (ours) & MV-NTF & LR-NTF-MVNTF (ours) \\
\cmidrule{3-8}          &       & \multicolumn{4}{c|}{Using endmembers extracted by VCA} & \multicolumn{2}{c}{Using endmembers extracted by MVNTF} \\
    \midrule
    \multirow{3}[2]{*}{SubDc1} & RE    & 0.0156  & 0.0154  & 0.0120  & \textbf{0.0099 } & 0.0194  & \textbf{0.0096 } \\
          & aSAM  & 0.1020  & 0.1015  & 0.0837  & \textbf{0.0623 } & 0.0980  & \textbf{0.0636 } \\
          & Time  & \textbf{15 } & 765   & 45    & 381   & 11664 & 1370 \\
    \midrule
    \multirow{3}[2]{*}{SubDc2} & RE    & 0.0094 & 0.0093  & 0.0080  & \textbf{0.0078 } & 0.0071  & \textbf{0.0027 } \\
          & aSAM  & 0.0264 & 0.0264  & 0.0246  & \textbf{0.0239 } & 0.0152  & \textbf{0.0076 } \\
          & Time  & \textbf{7 } & 636   & 20    & 218   & 3355  & 771 \\
    \bottomrule
    \end{tabular}%
    }
  \label{tab:addwash}%
\end{table*}%


\begin{figure}[htbp]
\centering
\includegraphics[scale=0.31]{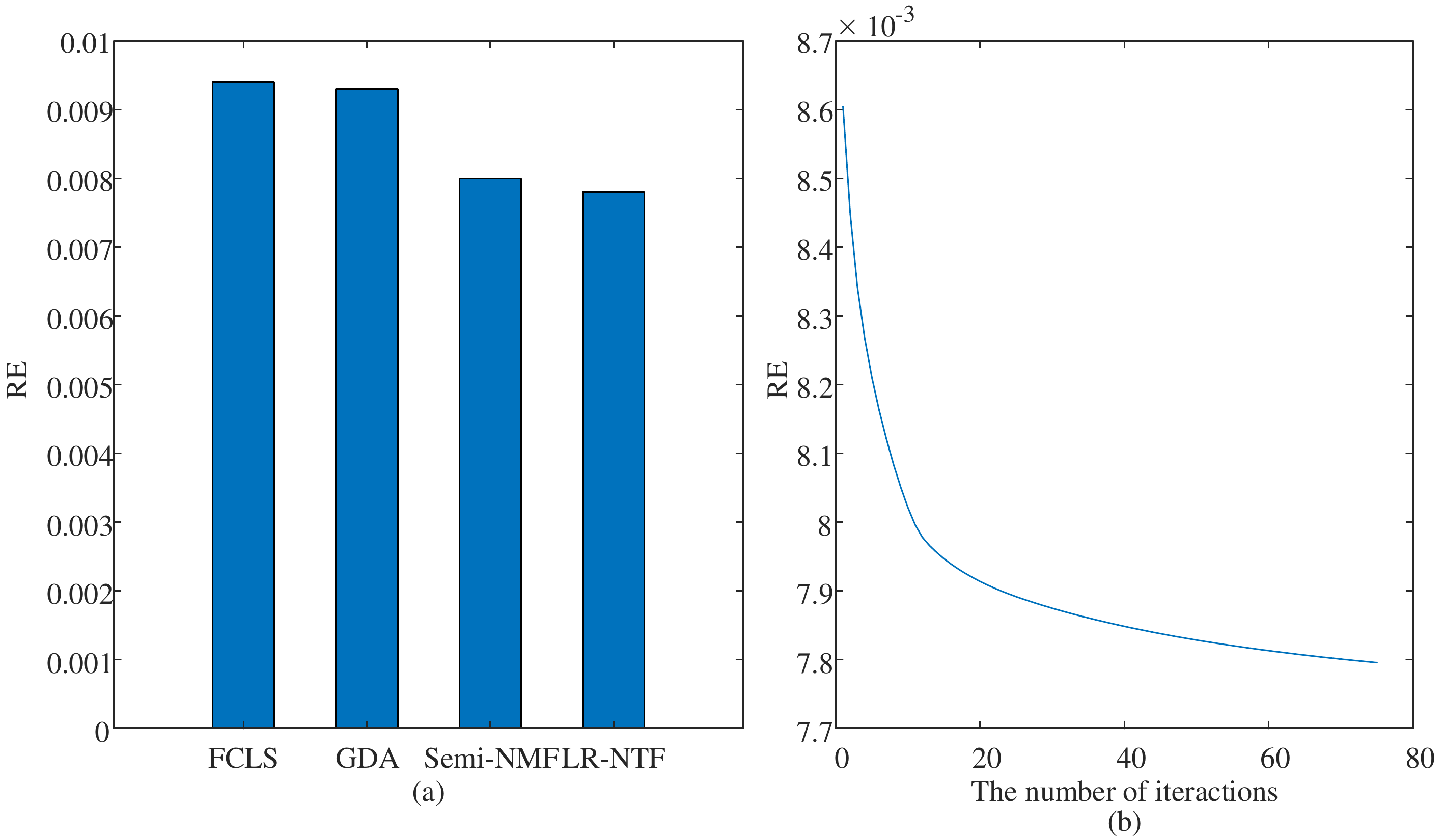}
\caption{RE of four unmixing methods in (a) and iteration of RE in the LR-NTF in (b) with the sub-image2 of  Washington DC Mall data}
\label{fig:re-wash2}
\end{figure}

\begin{figure}[htbp]
\centering
\includegraphics[scale=0.78]{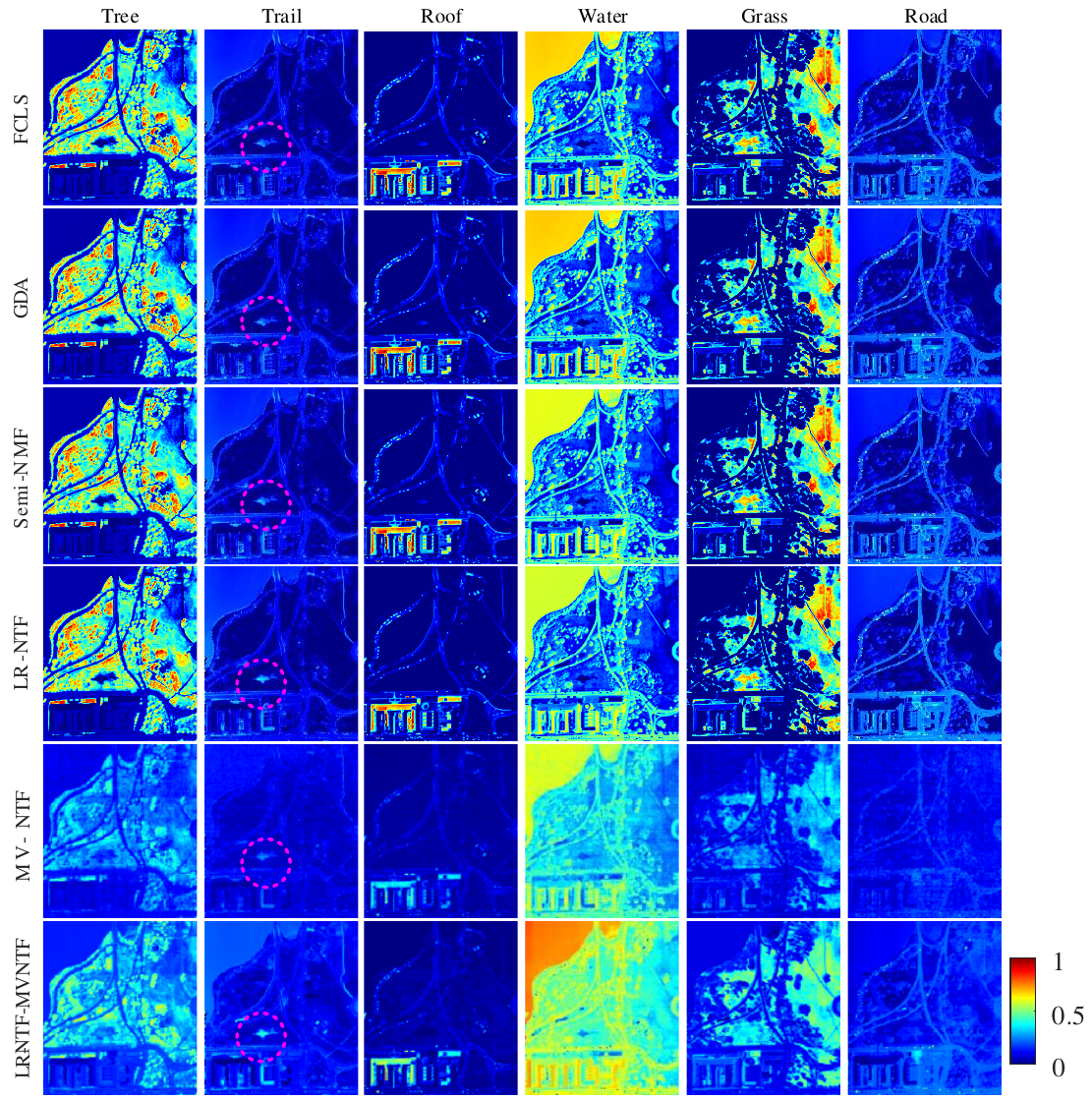}
\caption{Estimated abundance maps comparison between the proposed algorithm and state-of-the-art algorithms on sub-image2 of Washington DC Mall data}
\label{fig:abu-wash2}
\end{figure}

\begin{figure}[htbp]
\centering
\includegraphics[scale=0.28]{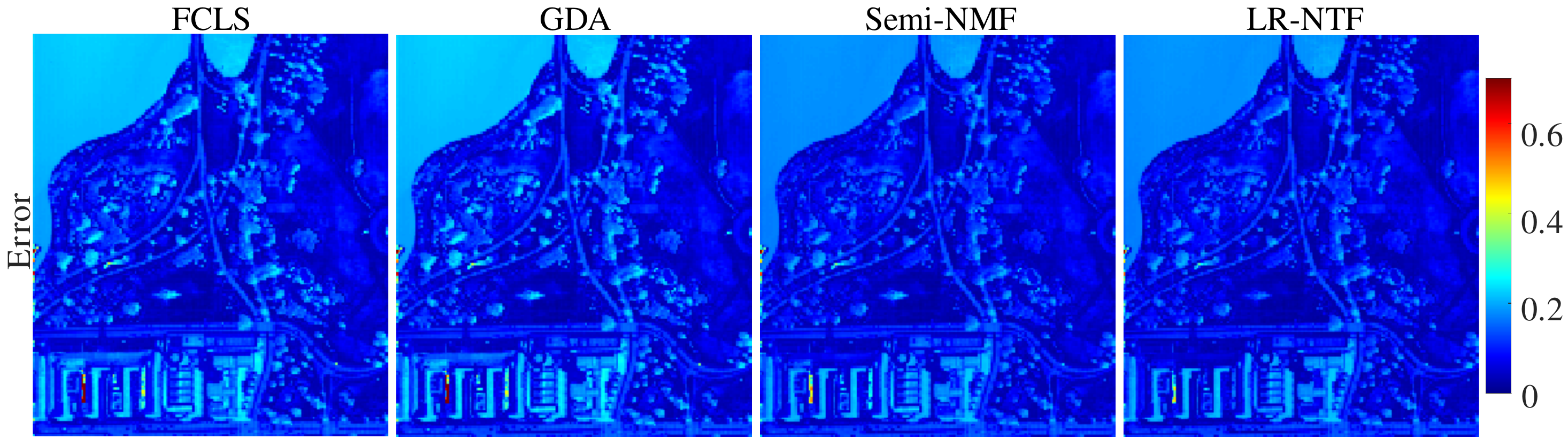}
\caption{RE distribution maps comparison between the proposed algorithm and state-of-the-art algorithms on sub-image2 of Washington DC Mall data}
\label{fig:err-wash2}
\end{figure}

The second sub-image, called sub-DC2, consisted of 180$\times$160 pixels and was also clipped from the Washington DC Mall data (see Fig. \ref{fig:realdata4}). This image has been studied in\cite{zhu2017spectral}, and six materials were extracted: Tree, Trail, Roof, Water, Grass, and Road. Hence, the number of endmembers, R, was set as 6 for our experiment; the endmember matrix was extracted using the VCA, and the parameter values used for all the algorithms were the same as for the first sub-image. Fig. \ref{fig:re-wash2}(a) shows that the proposed algorithm produces the lowest image reconstruction error and Fig. \ref{fig:abu-wash2} shows the estimated abundance maps. By taking into account the low-rank characteristics of the abundance maps and the third-order representation of the HSI, the proposed algorithm produces the best estimation of the abundance maps. Fig. \ref{fig:err-wash2} shows the error distribution for the whole image. The FCLS performs worst since this method fails to dig the complex information from the scene. The semi-NMF performs better than the GDA as it uses a matrix-based technique instead of a pixel-based unmixing method. Furthermore, the proposed algorithm produces a smaller error map than the FCLS, GDA, or semi-NMF.

In the experiments with  real images, we  have tested the proposed method  on the images with diverse spatial resolutions. Our method achieves good results in high spatial resolution images, including  Washington DC Mall with spatial resolution 3m  and San Diego Airport image with spatial resolution 3.5m. The spatial resolution of Cuprite image (20m) is relatively low, but our method still achieves good results.
A critical prior knowledge the proposed method used is the low-rankness of abundance maps and interaction abundance maps. The low-rankness exists because of the high spatial correlation of materials/endmbers, and does not depend on the spatial resolution of the image.
 Consequently, our proposed method could be used for images with diverse spatial resolutions, even lower spatial resolution than the ones discussed in the real data experiments.

\section{Conclusion}
In this paper, we proposed a nonnegative tensor factorization-based nonlinear hyperspectral unmixing method. By taking full advantage of the low rank of abundance maps, we imposed the nuclear norm on abundance maps and nonlinear interaction maps. In order to evaluate the effectiveness of the proposed algorithm, several kinds of  synthetic datasets and four real hyperspectral datasets were tested. Our method exploits the low-rank of abundance maps, and it was shown that this can improve the nonlinear unmixing performance. \cblue{Furthremore, we tested the proposed method in images with diverse number of endmembers (from 5 to 10), and obtained better results compared to other methods. However, with the
 growing  of the number of endmembers and the size of the images, the difficulty of unmixing will also increase, such as the need of more computation and memory, which is our future
research direction.}

\section*{Appendix A}

 \begin{table}
\centering
\setlength{\tabcolsep}{5mm}{
    \begin{tabular}{ll}
    \toprule
    \textbf{Abk.}  & \textbf{Bedeutung} \\
    \midrule
  aSAM       & \cblue{average of spectral angle mapper}          \\%
  BMMs       & \cblue{bilinear mixing models}          \\%
  GBM       & \cblue{generalized bilinear model}           \\%
  GDA       & \cblue{gradient descent algorithm}          \\%
  HNU       & \cblue{hyperspectral nonlinear unmixing}          \\%
  HSCs       & \cblue{hyperspectral cameras}          \\%
  HSIs       & \cblue{hyperspectral images}          \\%
  HU       & \cblue{hyperspectral unmixing}          \\%
  LMM       & \cblue{linear mixing model}          \\%
  LQM       & \cblue{linearquadratic mixing model}          \\%
  MHPNMM       & \cblue{multiharmonic postnonlinear mixing model } \\%\\
  MLM       & \cblue{multilinear mixing model}          \\%
  MTL       & \cblue{multi-task learning}          \\%
  NTF       & \cblue{nonnegative tensor factorization,}          \\%
  NLMMs       & \cblue{nonlinear mixing models}          \\%
  RE       & \cblue{reconstruction error}          \\%
  RMSE       & \cblue{root-mean-square error}          \\%
  semi-NMF       & \cblue{semi-nonnegative matrix factorization}          \\%
  VCA       & \cblue{vertex component analysis}          \\%
    \bottomrule
    \end{tabular}}%
  \label{tab:addlabel}%
\end{table}%

\section*{Acknowledgment}
The authors would like to thank Professor Naoto Yokoya for providing the semi-NMF code for our comparison experiment. Professor Yuntao Qian provided the endmember data used in some of the experiments with synthetic data.

\normalem
\bibliographystyle{IEEEtran}   
\bibliography{LR-NTF_FinalVersion}

\ifCLASSOPTIONcaptionsoff
  \newpage
\fi

\begin{IEEEbiography}
[{\includegraphics[width=1in,height=1.25in,clip,keepaspectratio]{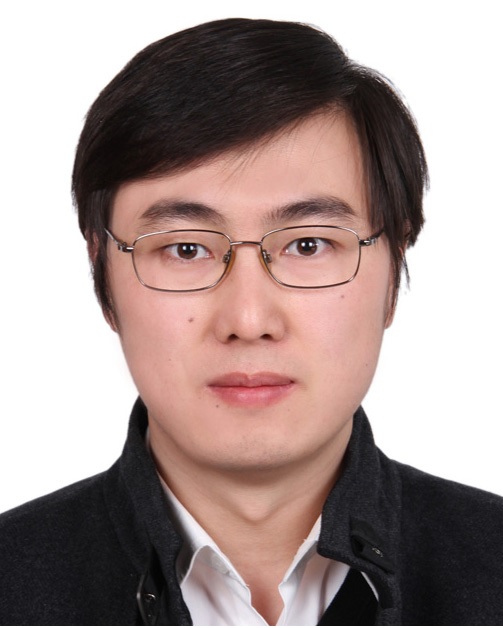}}]{Lianru Gao}
(Senior Member, IEEE) received the B.S. degree in civil engineering from Tsinghua University, Beijing, China, in 2002, and the Ph.D. degree in cartography and geographic information
system from the Institute of Remote Sensing
Applications, Chinese Academy of Sciences (CAS),
Beijing, in 2007.

He is a Professor with the Key Laboratory of Digital
Earth Science, Aerospace Information Research
Institute, CAS. He also has been a Visiting Scholar
with the University of Extremadura, C\'aceres, Spain,
in 2014, and Mississippi State University (MSU), Starkville, MS, USA, in 2016. In last 10 years, he was the principal investigator of ten scientific research projects at national and ministerial levels, including projects by the National Natural Science Foundation of China (2010-2012, 2016-2019, 2018-2020), and by the Key Research Program of the CAS (2013-2015) et al. He has authored or coauthored more than 160 peer-reviewed articles, and there are more than 80 journal articles included by Science Citation Index (SCI). He was a coauthor of an academic book \textit{Hyperspectral Image Classification and Target
Detection}. He holds 28 National Invention Patents in China. His research focuses on hyperspectral image processing and information extraction. Dr. Gao was awarded the Outstanding Science and Technology Achievement Prize of the CAS in 2016, and was supported by the China National Science Fund for Excellent Young Scholars in 2017, and won the Second Prize of The State Scientific and Technological Progress Award in 2018.
He received the recognition of the Best Reviewers of the IEEE JOURNAL OF SELECTED TOPICS IN APPLIED EARTH OBSERVATIONS AND REMOTE SENSING in 2015, and the Best Reviewers of the IEEE TRANSACTIONS ON GEOSCIENCE AND REMOTE SENSING in 2017.
\end{IEEEbiography}

\begin{IEEEbiography}
[{\includegraphics[width=1in,height=1.25in,clip,keepaspectratio]{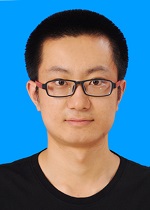}}]{Zhicheng Wang}
 received the B.E. degree  from the College of
Geosciences and Surveying Engineering, China University
of Mining and Technology (Beijing), Beijing, China, in 2018. He is pursuing the M.S. degree in electronics and communication  engineering with the Key Laboratory of Digital Earth Science, Aerospace Information Research Institute, Chinese Academy of Sciences (CAS), Beijing.

His research interests include image processing, machine learning, hyperspectral image nonlinear unmixing, and denoising.
\end{IEEEbiography}

\begin{IEEEbiography}
[{\includegraphics[width=1in,height=1.25in,clip,keepaspectratio]{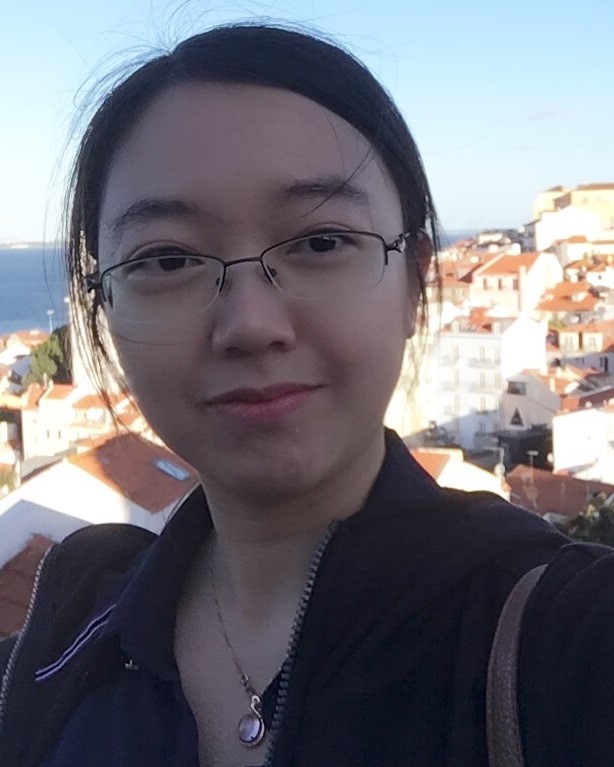}}]{Lina Zhuang}
(S'15-M'21) received Bachelor's  degrees in geographic information system and in economics from South China Normal University, Ghuangzhou, China,  in  2012,  the  M.S.  degree  in  cartography  and  geography  information  system  from  Institute  of  Remote  Sensing  and  Digital  Earth,  Chinese Academy of Sciences, Beijing, China, in 2015, and the Ph.D. degree in Electrical and Computer Engineering at the Instituto Superior Tecnico,  Universidade de Lisboa, Lisbon, Portugal in 2018. 

From 2015 to 2018, she was  a Marie Curie Early Stage Researcher of Sparse Representations and Compressed Sensing Training Network (SpaRTaN number 607290) with the Instituto de Telecomunica\c{c}\~{o}es. SpaRTaN Initial Training Networks (ITN) is funded under the European Union's Seventh Framework Programme  (FP7-PEOPLE-2013-ITN)  call  and  is  part  of  the  Marie  Curie Actions-ITN  funding  scheme. From 2019 to 2021, she was a Research Assistant Professor with    Hong Kong Baptist University.  She is currently  a Research Assistant Professor with the University of Hong Kong. Her research interests include hyperspectral image restoration, superresolution, and compressive sensing.
\end{IEEEbiography}

\begin{IEEEbiography}
[{\includegraphics[width=1in,height=1.25in,clip,keepaspectratio]{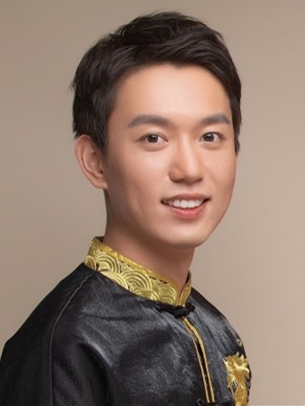}}]{Haoyang Yu}
(S’16-M’19) received the B.S. degree in Information and Computing Science from Northeastern University, Shenyang, China, in 2013. and the Ph.D. degree in cartography and geographic information system from the Key Laboratory of Digital Earth Science, Aerospace Information Research Institute, Chinese Academy of Sciences (CAS), Beijing, China, in 2019. 

He is currently a Xing Hai Associate Professor with the Center of Hyperspectral Imaging in Remote Sensing (CHIRS), Information Science and Technology College, Dalian Maritime University. His research focuses on models and algorithms for hyperspectral image processing, analysis and applications.
\end{IEEEbiography}

\begin{IEEEbiography}
[{\includegraphics[width=1in,height=1.25in,clip,keepaspectratio]{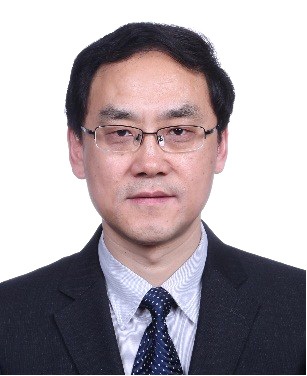}}]{Bing Zhang}
(Fellow, IEEE) received the B.S.
degree in geography from Peking University, Beijing, China, in 1991, and the M.S. and Ph.D. degrees in remote sensing from the Institute of Remote Sensing Applications, Chinese Academy
of Sciences (CAS), Beijing, in 1994 and 2003, respectively.

He is a Full Professor and the Deputy Director of the Aerospace Information Research Institute, CAS, where he has been leading lots of key scientific projects in the area of hyperspectral remote sensing for more than 25 years. He has developed five software systems in the image processing and applications. His creative achievements were rewarded ten important prizes from the Chinese government, and special government allowances of the Chinese State Council. He has authored more than 300 publications, including more than 170 journal articles. He has edited six books/contributed book chapters on hyperspectral image processing and subsequent applications. His research interests include the development of Mathematical and Physical models and image processing software for the analysis of hyperspectral remote sensing data in many different areas.

Dr. Zhang has been serving as a Technical Committee Member of IEEE Workshop on Hyperspectral Image and Signal Processing, since 2011, and has been the President of Hyperspectral Remote Sensing Committee of China National Committee of International Society for Digital Earth since 2012, and has been the Standing Director of Chinese Society of Space Research (CSSR) since 2016. He is the Student Paper Competition Committee Member in International Geoscience and Remote Sensing Symposium from 2015 to 2020. He was awarded the National Science Foundation for Distinguished Young Scholars of China in 2013, and the 2016 Outstanding Science and Technology Achievement Prize of the Chinese Academy of Sciences, the highest level of Awards for the CAS scholars. He is serving as an Associate Editor for the IEEE JOURNAL OF SELECTED TOPICS IN APPLIED EARTH OBSERVATIONS AND REMOTE SENSING.
\end{IEEEbiography}

\begin{IEEEbiography}
[{\includegraphics[width=1in,height=1.25in,clip,keepaspectratio]{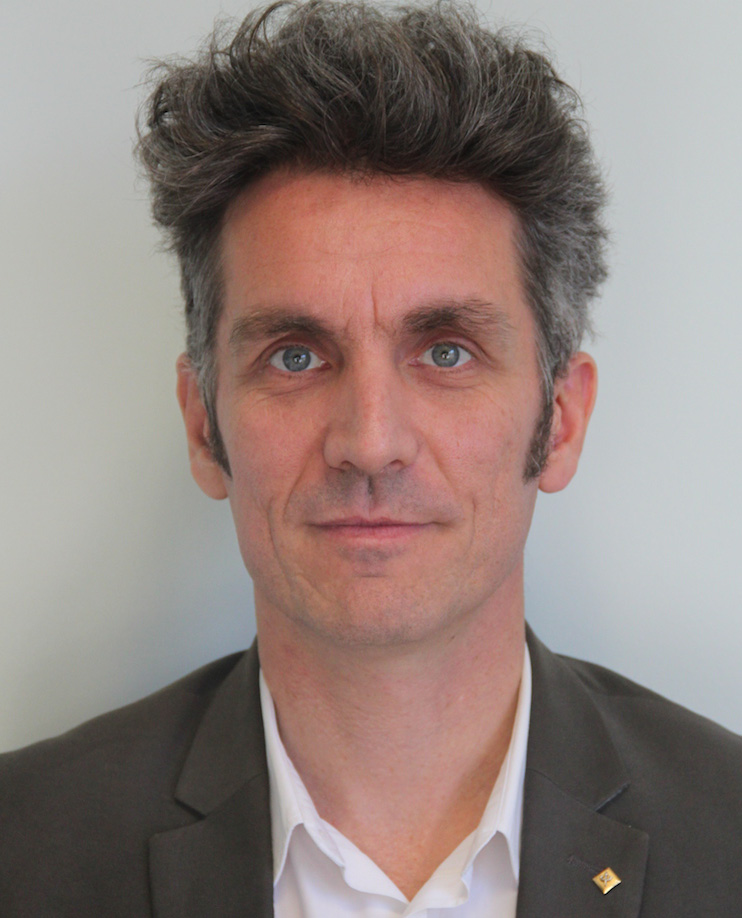}}]{Jocelyn Chanussot}
(Fellow, IEEE) received the
M.Sc. degree in electrical engineering from the
Grenoble Institute of Technology (Grenoble INP),
Grenoble, France, in 1995, and the Ph.D. degree
from the Université de Savoie, Annecy, France,
in 1998.

Since 1999, he has been with Grenoble INP, where
he is a Professor of signal and image processing. His
research interests include image analysis, hyperspectral
remote sensing, data fusion, machine learning
and artificial intelligence. He has been a Visiting
Scholar with Stanford University, Stanford, CA, USA, KTH Royal Institute
of Technology, Stockholm, Sweden, and National University of Singapore,
Singapore. Since 2013, he is an Adjunct Professor with the University of
Iceland, Reykjavík, Iceland. In 2015–2017, he was a Visiting Professor with
the University of California, Los Angeles (UCLA), Los Angeles, CA, USA.
He holds the AXA chair in remote sensing and is an Adjunct Professor with
the Chinese Academy of Sciences, Aerospace Information Research Institute,
Beijing.

Dr. Chanussot is the founding President of IEEE Geoscience and Remote
Sensing French chapter (2007–2010) which received the 2010 IEEE Geoscience and Remote Sensing Society
Chapter Excellence Award. He has received multiple outstanding paper
awards. He was the Vice-President of the IEEE Geoscience and Remote
Sensing Society, in charge of meetings and symposia (2017–2019). He was
the General Chair of the first IEEE GRSS Workshop on Hyperspectral Image
and Signal Processing, Evolution in Remote sensing (WHISPERS). He was
the Chair (2009–2011) and Cochair of the GRS Data Fusion Technical
Committee (2005–2008). He was a Member of the Machine Learning for
Signal Processing Technical Committee of the IEEE Signal Processing Society
(2006–2008) and the Program Chair of the IEEE International Workshop on
Machine Learning for Signal Processing (2009). He is an Associate Editor
for the IEEE GEOSCIENCE AND REMOTE SENSING LETTERS, the IEEE
TRANSACTIONS ON IMAGE PROCESSING, and the PROCEEDINGS OF THE
IEEE. He was the Editor-in-Chief of the IEEE JOURNAL OF SELECTED
TOPICS IN APPLIED EARTH OBSERVATIONS AND REMOTE SENSING (2011–
2015). In 2014, he served as a Guest Editor for the IEEE Signal Processing
Magazine. He is a Member of the Institut Universitaire de France (2012–
2017) and a Highly Cited Researcher (Clarivate Analytics/Thomson Reuters,
2018–2019).
\end{IEEEbiography}

\end{document}